\documentclass{article}

\usepackage{microtype}
\usepackage{graphicx}
\usepackage{subfigure}
\usepackage{booktabs} 
\usepackage{amsmath}
\usepackage{listings}
\usepackage{xcolor}
\usepackage{hyperref}
\usepackage{lineno}

\definecolor{codegreen}{rgb}{0,0.6,0}
\definecolor{codegray}{rgb}{0.5,0.5,0.5}
\definecolor{codepurple}{rgb}{0.58,0,0.82}
\definecolor{backcolour}{rgb}{0.95,0.95,0.92}
\definecolor{lightgray}{gray}{0.95}

\lstdefinestyle{mystyle}{
    backgroundcolor=\color{lightgray},   
    commentstyle=\color{codegreen},
    keywordstyle=\color{magenta},
    numberstyle=\tiny\color{codegray},
    stringstyle=\color{codepurple},
    basicstyle=\ttfamily\footnotesize,
    breakatwhitespace=false,         
    breaklines=true,                 
    captionpos=b,                    
    keepspaces=true,                 
    numbers=left,                    
    numbersep=5pt,                  
    showspaces=false,                
    showstringspaces=false,
    showtabs=false,                  
    tabsize=2
}

\usepackage[accepted]{mlsys2025}

\makeatletter
\newcommand{\mlsys@appearing}{}
\makeatother

\mlsystitlerunning{Lossless Prompt Compression via Dictionary-Encoding and In-Context Learning}

\begin{document}

\twocolumn[
\mlsystitle{Lossless Prompt Compression via Dictionary-Encoding and In-Context Learning: Enabling Cost-Effective LLM Analysis of Repetitive Data}



\mlsyssetsymbol{equal}{*}

\begin{mlsysauthorlist}
\mlsysauthor{Andresa Rodrigues de Campos}{amz}
\mlsysauthor{David Lee}{amz}
\mlsysauthor{Imry Kissos}{amz}
\mlsysauthor{Piyush Paritosh}{amz}
\end{mlsysauthorlist}

\mlsysaffiliation{amz}{Amazon.com}

\mlsyscorrespondingauthor{Andresa Rodrigues de Campos}{ardc@amazon.com}

\mlsyskeywords{}

\vskip 0.3in

\begin{abstract}

In-context learning has established itself as an important learning paradigm for Large Language Models (LLMs). In this paper, we demonstrate that LLMs can learn encoding keys in-context and perform analysis directly on encoded representations. This finding enables lossless prompt compression via dictionary encoding without model fine-tuning: frequently occurring subsequences are replaced with compact meta-tokens, and when provided with the compression dictionary in the system prompt, LLMs correctly interpret these meta-tokens during analysis, producing outputs equivalent to those from uncompressed inputs. We present a compression algorithm that identifies repetitive patterns at multiple length scales, incorporating a token-savings optimization criterion that ensures compression reduces costs by preventing dictionary overhead from exceeding savings. The algorithm achieves compression ratios up to 80\% depending on dataset characteristics. To validate that LLM analytical accuracy is preserved under compression, we use decompression as a proxy task with unambiguous ground truth. Evaluation on the LogHub 2.0 benchmark using Claude 3.7 Sonnet demonstrates exact match rates exceeding 0.99 for template-based compression and average Levenshtein similarity scores above 0.91 for algorithmic compression, even at compression ratios of 60\%–80\%. Additionally, compression ratio explains less than 2\% of variance in similarity metrics, indicating that decompression quality depends on dataset characteristics rather than compression intensity. This training-free approach works with API-based LLMs, directly addressing fundamental deployment constraints --- token limits and API costs --- and enabling cost-effective analysis of large-scale repetitive datasets, even as data patterns evolve over time.

\end{abstract}

]
\printAffiliationsAndNotice{}

\section{Introduction}
\label{sec:introduction}

Large Language Models (LLMs) face significant cost and latency challenges when processing repetitive textual data due to token-based pricing and context window constraints. Enterprise applications routinely generate logs with thousands of repetitive entries, e.g., identical patterns, timestamps, and system paths repeated across execution traces. Processing such data through LLMs
for tasks like system monitoring (error traces, performance metrics) or process mining (event sequences), for example, 
incurs substantial costs that scale directly with token consumption, creating barriers to comprehensive analysis of these large-scale datasets.

This motivates us to explore lossless compression as a token reduction strategy. Traditional compression algorithms achieve high ratios on repetitive data, but LLMs cannot natively process compressed formats. The critical question becomes: Can LLMs analyze compressed data directly without decompression? If LLMs could learn the encoding key in-context and perform analytical reasoning on encoded representations, we could achieve substantial cost reduction without sacrificing analytical fidelity.

This insight leads us to reexamine dictionary-encoding compression \cite{Salomon:2004:DCC} from an LLM-centric perspective, focusing on whether in-context learning capabilities enable LLMs to internalize compression mappings and produce equivalent analytical outputs from compressed inputs. System logs provide an ideal testbed for this paradigm, as they involve highly repetitive data with well-defined analytical objectives and quantitative evaluation metrics.

Accordingly, we present a hierarchical dictionary-encoding algorithm designed for LLM-based analysis. Our work makes three primary contributions:

First, we demonstrate empirically that LLMs can learn compression dictionaries in-context and produce analytically equivalent outputs from compressed data. Validation experiments on LogHub~2.0 system logs~\cite{zhu2023loghub,jiang2024loghub2} achieve 0.99+ exact match rates for template-based compression and similarity scores above 0.91 on average for algorithmic compression, even at compression ratios of 60\%-80\%. LLM analytical performance remains robust across all compression levels, challenging conventional assumptions about compression trade-offs. This finding eliminates the need for model fine-tuning and enables deployment with standard API-based LLMs.

Second, we develop a hierarchical processing strategy that identifies repetitive subsequences at multiple length scales and applies replacements in descending order of token savings. This approach handles nested patterns, prevents overlapping replacements, and achieves deterministic compression suitable for production deployment. The algorithm integrates seamlessly with batch processing architectures, generating optimized dictionaries per batch to maximize compression efficiency.

Third, we introduce a token-savings optimization criterion that ensures compression reduces costs (see Section \ref{sec:sec:savins_cond}). This criterion prevents dictionary overhead from exceeding savings and adapts compression aggressiveness to pattern characteristics, maximizing cost reduction while maintaining analytical equivalence.

In summary, this work presents a practical approach to reducing LLM costs for repetitive data analysis through lossless compression with in-context dictionary learning. Section~\ref{sec:background} presents the current state of the field and related work. Section~\ref{sec:method} formalizes the problem and presents our hierarchical compression algorithm with the token-savings optimization criterion. Section~\ref{sec:exp} describes the experimental setup using the LogHub 2.0 dataset, including both template-based and algorithmic compression evaluations. Section~\ref{sec:results} presents compression efficiency across datasets and validates LLM decompression capabilities. Section~\ref{sec:discussion} analyzes implications for cost-effective LLM deployment and application domains. Section~\ref{sec:limitations} addresses limitations and future directions, while Section~\ref{sec:conclusion} summarizes key contributions.

\section{Background and Related Work}
\label{sec:background}

Recent work on prompt compression for LLMs falls into \textit{lossy} and \textit{lossless} categories. 
Lossy strategies compress prompts with minimal information loss. The \textit{In-Context Former} proposed by Zhang et al.~\cite{zhang2024incontext} trains a compact encoder to create ``digest'' tokens, achieving 4$\times$ compression and over 90\% retention in downstream performance. In contrast, methods such as Hou et al.~\cite{hou2024instruction} use instruction-aware filtering to remove irrelevant text, improving efficiency with no training overhead.

LLMLingua~\cite{jiang2023llmlingua} uses a small language model to identify and remove non-essential tokens based on perplexity, achieving up to 20$\times$ compression while maintaining performance on reasoning tasks. LLMLingua-2~\cite{pan2024llmlingua2} extends this approach through data distillation, training a task-agnostic compressor that preserves essential information more faithfully.

When training is not feasible, heuristics-based methods offer strong performance. Xu et al.~\cite{xu2024xraglog} propose \textit{XRAGLog}, a compression approach tailored for logs that leverages structured templates and retrieval to compress up to 90\% of log tokens while improving anomaly detection performance.

For compressing LLM \textit{outputs}, lossless methods such as sequence modeling have emerged. Mao et al.~\cite{mao2024llmzip} and Narasimhan \& Chandrachoodan~\cite{narasimhan2024alphazip} show that transformer-based models can predict token distributions and apply entropy coding to achieve up to 20$\times$ compression over baseline methods.

Harvill et al.~\cite{harvill2025lossless} introduced the first lossless strategy compressing LLM \textit{inputs}. It replaces repeated token subsequences with \textit{meta-tokens}, achieving 18--27\% token reduction with corresponding 33--47\% compute savings while maintaining recoverability. However, their approach requires fine-tuning the LLM to understand placeholder token meanings, creating a fundamental scalability limitation. 

Log data is particularly well-suited for compression due to its regularity. \textit{LogHub}~\cite{zhu2023loghub, jiang2024loghub2} compile a diverse benchmark of real-world structured logs. Methods such as LogShrink~\cite{li2023logshrink} exploit recurring patterns in logs for high compression ratios. These datasets are suitable for benchmarking compression accuracy, efficiency, and recovery fidelity.

Metrics commonly used to evaluate context compression include compression ratio, task accuracy on decompressed inputs, and exact recovery rate. Lossless methods such as meta-token substitution~\cite{harvill2025lossless} or XRAGLog-style grouping~\cite{xu2024xraglog} are favored for high-stakes log analysis, while lossy methods enable broader reductions where some information loss is tolerable.

Our work focuses on compressing LLM \textit{input} data and addresses a critical gap in existing approaches. While our approach shares the dictionary-based compression concept with methods like~\cite{harvill2025lossless}, our methodology is fundamentally different as it leverages the LLM's inherent natural language understanding without requiring any model modifications. This training-free approach provides immediate applicability, making it particularly suitable for production environments where data characteristics change over time and retraining costs would be prohibitive. By maintaining high compression ratios while preserving decompression fidelity, our method offers a practical solution for large-scale log analysis without the operational overhead of specialized model training.

\section{Methodology}
\label{sec:method}

\begin{figure*}[h!]
    \centering
    \includegraphics[width=2.0\columnwidth]{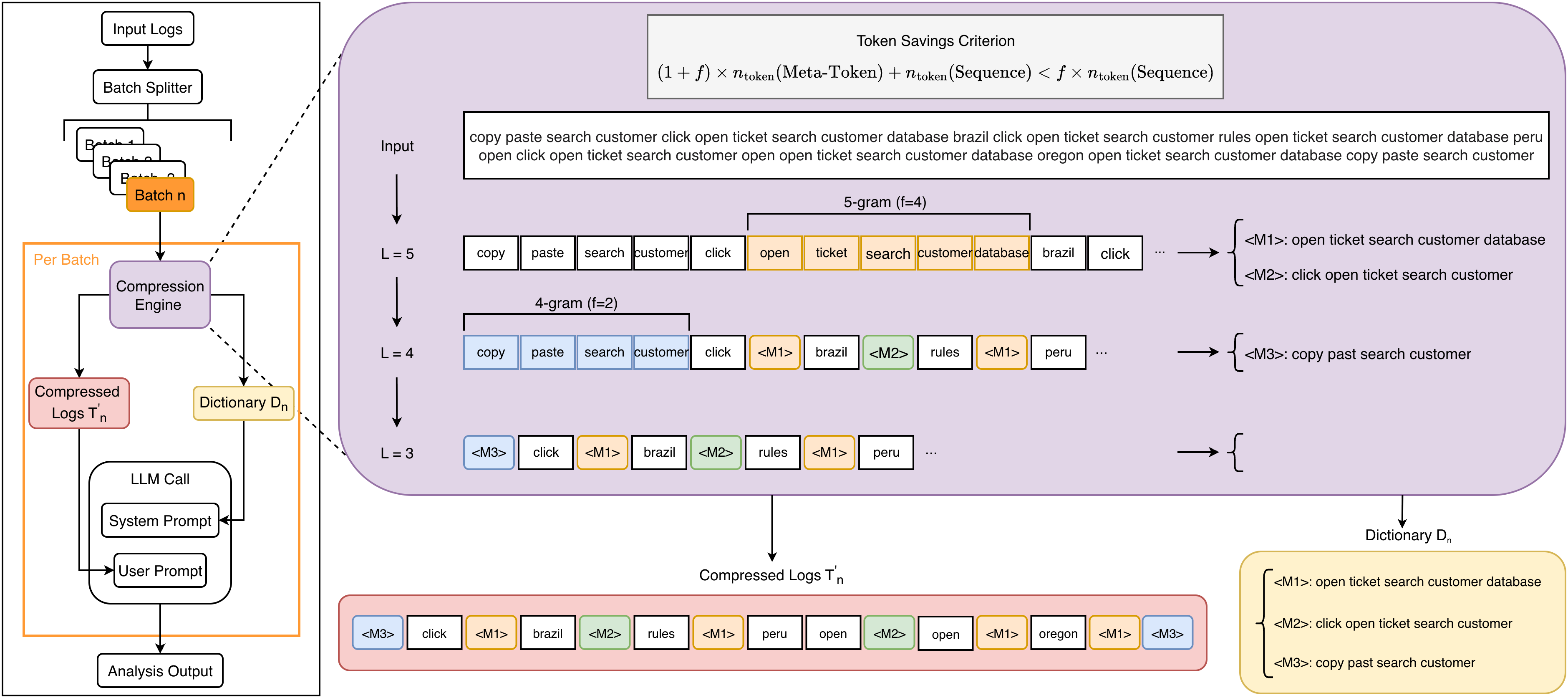}
    \caption{System overview showing hierarchical compression algorithm (right), subsequence finding detail (top left), and compressed output with dictionary (bottom left).}
    \label{fig:system}
\end{figure*}

\subsection{Problem Formulation}

Given input text $T$ containing repetitive subsequences, our goal is to generate compressed text $T'$ and dictionary $\mathcal{D}$ such that: (1)~$T$ can be perfectly reconstructed from $T'$ and $\mathcal{D}$ (lossless reconstruction), (2)~$n_{\text{token}}(T' + \mathcal{D}) < n_{\text{token}}(T)$ (token reduction), and (3)~LLM analysis of $(T', \mathcal{D})$ produces outputs equivalent to analysis of $T$ (analytical equivalence).

The challenge lies in identifying which subsequences to compress such that dictionary overhead is offset by compression gains, while ensuring the LLM can interpret the compressed representation.

\subsection{Dictionary-Encoding Compression}

Our compression approach implements dictionary-encoding through four stages: 

(1)~\textit{Whitespace segmentation}: The input text is segmented at whitespace boundaries to produce a sequence of non-empty word units.
(2)~\textit{Subsequence Identification}: Algorithm~\ref{alg:hierarchical_compression} identifies frequently occurring subsequences through hierarchical processing. Algorithm~\ref{alg:find_subsequences} evaluates which subsequences to keep. 
(3)~\textit{Replacement Application}: Algorithm~\ref{alg:apply_replacements} replaces selected subsequences with meta-tokens in format $\langle M\# \rangle$ (e.g., $\langle M1 \rangle$, $\langle M2 \rangle$). 
(4)~\textit{Dictionary Construction}: The final dictionary maps meta-tokens to original subsequences, allowing lossless reconstruction.

\subsection{Compression Algorithm}\label{sec:sec:comp_algo} 

\begin{algorithm}[tb]
   \caption{Hierarchical Compression}
   \label{alg:hierarchical_compression}
\begin{algorithmic}
   \STATE {\bfseries Input:} token sequence $T$, max length $L_{max}$, min frequency $f_{min}$
   \STATE {\bfseries Output:} compressed sequence $T'$, dictionary $D$
   \STATE Initialize $T' \leftarrow T$, $D \leftarrow \emptyset$, $meta\_counter \leftarrow 1$
   \FOR{$\ell = L_{max}$ {\bfseries down to} $2$}
   \STATE $S \leftarrow \emptyset$ \COMMENT{Selected subsequences}
   \STATE $used\_pos \leftarrow \emptyset$ \COMMENT{Used positions}
   \FOR{each subsequence $s$ of length $\ell$ in $T'$}
   \IF{$s$ contains no meta-tokens {\bfseries and} frequency$(s) \geq f_{min}$}
   \STATE Find non-overlapping positions $P$ for $s$
   \IF{$|P| \geq f_{min}$ {\bfseries and} $(|P|+1) \cdot ntoken(meta) + ntoken(s) < |P| \cdot ntoken(s)$}
   \STATE $meta \leftarrow \langle M_{meta\_counter} \rangle$
   \STATE $S \leftarrow S \cup \{(s, P, meta)\}$
   \STATE $used\_pos \leftarrow used\_pos \cup \bigcup_{p \in P} \{p, p+1, \ldots, p+\ell-1\}$
   \STATE $meta\_counter \leftarrow meta\_counter + 1$
   \ENDIF
   \ENDIF
   \ENDFOR
   \FOR{each $(s, P, meta) \in S$}
   \STATE Replace all occurrences of $s$ at positions $P$ with $meta$ in $T'$
   \STATE $D[meta] \leftarrow s$
   \ENDFOR
   \ENDFOR
   \STATE \textbf{return} $T'$, $D$
\end{algorithmic}
\end{algorithm}

Algorithm~\ref{alg:hierarchical_compression} takes three parameters: the token sequence $T$, maximum subsequence length $L_{max}$, and minimum frequency threshold $f_{min}$. The parameter $L_{max}$ controls the upper bound on subsequence length considered for compression. The algorithm iterates through lengths $\ell$ from $L_{max}$ down to a minimum length (typically 2). Higher $L_{max}$ values allow compression of longer repetitive patterns, potentially achieving greater compression at the cost of increased computational complexity. The algorithm processes subsequences hierarchically from longest to shortest, ensuring longer patterns are compressed before shorter ones to maximize compression efficiency. 


For each length, Algorithm~\ref{alg:find_subsequences} performs the core subsequence identification: (1)~\textit{Subsequence Scanning}: Extract all subsequences of length $\ell$, skipping those containing existing meta-tokens to prevent nested compression. (2)~\textit{Frequency Analysis}: Group identical subsequences and filter to those with frequency $\geq f_{\min}$. (3)~\textit{Non-overlapping Selection}: Sort by frequency (descending) and select subsequences with non-overlapping positions. (4)~\textit{Token Savings Evaluation}: Apply the token savings condition (see Section \ref{sec:sec:savins_cond}) to determine compression benefit. Selected subsequences are then processed by Algorithm~\ref{alg:apply_replacements} to generate the updated token sequence.

The algorithm maintains a working token sequence that is incrementally updated during hierarchical processing, ensuring subsequent iterations operate on partially compressed data.
Figure~\ref{fig:system} illustrates the complete compression system, showing the system flow (right), detailed hierarchical subsequence finding process (top left), and resulting compressed output with dictionary (bottom left).

\begin{algorithm}[tb]
   \caption{Find Subsequences at Length}
   \label{alg:find_subsequences}
\begin{algorithmic}
   \STATE {\bfseries Input:} token sequence $T$, length $\ell$, min frequency $f_{min}$, meta counter
   \STATE {\bfseries Output:} selected subsequences with positions and meta-tokens
   \STATE Initialize $subseq\_pos \leftarrow \emptyset$, $used\_pos \leftarrow \emptyset$
   \FOR{$i = 0$ {\bfseries to} $|T| - \ell$}
   \STATE $s \leftarrow T[i:i+\ell]$
   \IF{$s$ contains no meta-tokens}
   \STATE $subseq\_pos[s] \leftarrow subseq\_pos[s] \cup \{i\}$
   \ENDIF
   \ENDFOR
   \STATE Filter $subseq\_pos$ to keep only entries with $|positions| \geq f_{min}$
   \STATE Sort by frequency (descending)
   \STATE $selected \leftarrow \emptyset$
   \FOR{each $(s, positions) \in subseq\_pos$}
   \STATE $valid\_pos \leftarrow \{p \in positions : [p, p+\ell) \cap used\_pos = \emptyset\}$
   \IF{$|valid\_pos| \geq f_{min}$ {\bfseries and} token savings condition holds}
   \STATE $meta \leftarrow \langle M_{counter} \rangle$, increment $counter$
   \STATE $selected \leftarrow selected \cup \{(s, valid\_pos, meta)\}$
   \STATE $used\_pos \leftarrow used\_pos \cup \bigcup_{p \in valid\_pos} [p, p+\ell)$
   \ENDIF
   \ENDFOR
   \STATE \textbf{return} $selected$
\end{algorithmic}
\end{algorithm}

\subsection{Token Savings Condition}\label{sec:sec:savins_cond}

The token savings condition ensures compression provides net benefit by comparing token cost of compressed versus uncompressed representations. A subsequence $S$ appearing $f$ times costs $f \cdot n_{\text{token}}(S)$ tokens uncompressed, where $n_{\text{token}}(\cdot)$ is a function that computes the tokens cost. Using meta-token $M$ to replace $S$ costs: dictionary entry $n_{\text{token}}(S)$ tokens, dictionary label $n_{\text{token}}(M)$ tokens, and meta-token occurrences $f \cdot n_{\text{token}}(M)$ tokens, totaling $(1 + f) \cdot n_{\text{token}}(M) + n_{\text{token}}(S)$ tokens compressed.

Compression provides benefit when:
\begin{equation}
    (1 + f) \cdot n_{\text{token}}(M) + n_{\text{token}}(S) < f \cdot n_{\text{token}}(S)
    \label{eq:token_savings}
\end{equation}

This condition is evaluated in Algorithm~\ref{alg:find_subsequences} for each candidate subsequence. Only subsequences satisfying the condition are compressed.


\subsection{Overlap Prevention}

Algorithm~\ref{alg:find_subsequences} maintains a set of used positions to prevent overlapping compressions, ensuring each token is compressed at most once. When a subsequence at positions $[i, i+\ell)$ is selected, all positions $\{i, i+1, \ldots, i+\ell-1\}$ are marked as used. During evaluation of subsequent candidates, positions overlapping with used positions are filtered out before frequency recomputation.

Algorithm~\ref{alg:apply_replacements} implements the replacement strategy by building a replacement map and skip positions set, ensuring clean non-overlapping substitutions. The algorithm also prevents nested compressions by checking if a subsequence contains existing meta-tokens during the scanning phase.

\begin{algorithm}[tb]
   \caption{Apply Replacements}
   \label{alg:apply_replacements}
\begin{algorithmic}
   \STATE {\bfseries Input:} token sequence $T$, selected subsequences $S$
   \STATE {\bfseries Output:} modified sequence $T'$
   \STATE Initialize $replacement\_map \leftarrow \emptyset$, $skip\_pos \leftarrow \emptyset$
   \FOR{each $(s, positions, meta) \in S$}
   \FOR{each $p \in positions$}
   \STATE $replacement\_map[p] \leftarrow meta$
   \STATE $skip\_pos \leftarrow skip\_pos \cup \{p+1, p+2, \ldots, p+|s|-1\}$
   \ENDFOR
   \ENDFOR
   \STATE Initialize $T' \leftarrow []$
   \FOR{$i = 0$ {\bfseries to} $|T|-1$}
   \IF{$i \in replacement\_map$}
   \STATE $T' \leftarrow T' \cup \{replacement\_map[i]\}$
   \ELSIF{$i \notin skip\_pos$}
   \STATE $T' \leftarrow T' \cup \{T[i]\}$
   \ENDIF
   \ENDFOR
   \STATE \textbf{return} $T'$
\end{algorithmic}
\end{algorithm}

\subsection{Batch-Level Compression Strategy}

For large datasets, compression is applied at the batch level, with each batch receiving a batch-specific dictionary optimized for patterns within that batch.
%
As shown in Figure~\ref{fig:system}, data is divided into batches that fit into the LLM's context window. For each batch: (1)~Apply compression algorithm independently. (2)~Generate batch-specific dictionary containing only patterns present in that batch. (3)~Construct LLM user prompt with compressed batch and system prompt with batch-specific dictionary. (4)~Process batch through LLM. (5)~Aggregate results across batches.

Benefits include: meta-tokens are not wasted on sequences absent from the batch, batches can be compressed and processed independently (enabling parallelization), dictionary size is proportional to batch-specific patterns (not global patterns), and each batch receives a dictionary optimized for its specific repetition characteristics.

\section{Experiments}
\label{sec:exp}

We utilize the LogHub 2.0 dataset~\cite{zhu2023loghub}, a comprehensive collection of system logs for AI-driven log analytics research. The dataset includes logs from distributed systems (Hadoop, HDFS, OpenStack, Spark, Zookeeper), supercomputer systems (BGL, HPC, Thunderbird), operating systems (Linux, Mac), server applications (Apache, OpenSSH), and standalone software (HealthApp, Proxifier).


The dataset provides two scales: LogHub-2k contains 2,000 annotated logs per system, while the full LogHub-2.0 contains complete datasets ranging from 21,320 logs (Proxifier) to 16.6 million logs (Thunderbird). It also includes log templates with counts varying from 6 (Apache-2k) to 1,241 (Thunderbird-full). We conduct our primary experiments on the LogHub-2k datasets, which range from approximately 14,000 to 111,000 tokens depending on the system (see Appendix~\ref{appendix:token_stats}), and validate scalability on selected full datasets.

The dataset is particularly suitable for validation due to: (1)~High repetition: system logs contain numerous repeated patterns (timestamps, error codes, system paths). (2)~Structural consistency: logs follow predictable formats ideal for dictionary encoding. (3)~Domain diversity: includes logs from multiple system types. (4)~Real-world data: logs represent authentic production scenarios.

\subsection{Compression Efficiency}

First, we establish the compression feasibility of each dataset using our hierarchical compression algorithm. This analysis characterizes the intrinsic compressibility of different log types and informs expectations for subsequent experiments.

We define compression ratio ($CR$) as:

\begin{equation}
    CR = 1 - \frac{n_{\text{token}}(\text{Compressed Data})}{n_{\text{token}}(\text{Original Data})}
\end{equation}

\noindent where $n_{\text{token}}(\cdot)$ denotes the token count. This metric can be computed in two ways depending on the analytical context:

\textbf{Compressed logs only}: When measuring intrinsic data compressibility, we compute $CR$ using only the compressed log tokens. This reveals how much redundancy exists in the original data and represents an upper bound on achievable compression.

\textbf{Compressed logs + dictionary}: When measuring the actual token reduction for LLM input, we must account for the dictionary overhead since both the compressed data and the dictionary are provided to the model. This represents the true compression ratio relevant for cost savings:

\begin{equation}
    CR_{\text{input}} = 1 - \frac{n_{\text{token}}(\text{Compressed Data}) + n_{\text{token}}(\text{Dict})}{n_{\text{token}}(\text{Original Data})}
\end{equation}

The distinction is important: intrinsic data compressibility does not guarantee cost savings if the dictionary overhead is substantial. Equation~\ref{eq:token_savings} ensures that compression is applied only when $CR_{\text{input}} > 0$, but the magnitude of savings depends on the balance between pattern frequency and dictionary size.

We measure compression ratios for each dataset across $L_{max}$ values from 3 to 20, evaluating both metrics to understand the relationship between intrinsic compressibility and practical token reduction.

\subsection{Decompression Validation}
\label{sec:sec:decompression}

We validate whether LLMs can effectively decompress compressed data. This serves as a proxy for the LLM's ability to understand compressed data in-context and analyze it. Decompression is well-suited for this evaluation because it offers unambiguous ground truth and objective evaluation metrics. It is important to notice, however, that decompression adds an extra challenge to understanding: an LLM that fully understands the compressed data may still produce minor reconstruction errors during decompression given that reconstruction is itself an additional generative task with its own error modes. Thus, our decompression metrics represent a lower bound on the LLM's true comprehension of the compressed input. 


We conduct two complementary experiments: template-based compression using existing log templates to assess LLM decompression capabilities in Section~\ref{sec:sec:sec:template-decompression}, and end-to-end evaluation using our hierarchical algorithm in Section~\ref{sec:sec:sec:algo-decompression}.

\subsubsection{Template-Based Decompression}\label{sec:sec:sec:template-decompression}

\textbf{Experimental Design}

We first validate LLM decompression capabilities using existing log templates from the LogHub dataset to create dictionaries that map meta-tokens to template patterns. This provides a controlled evaluation environment where the compression dictionary is derived from known log structures rather than our algorithm. 

The LLM receives a system prompt containing analysis instructions and the compression dictionary, with compressed log data containing meta-tokens in the user prompt.
The LLM is instructed to decode the data by replacing each meta-token with its corresponding sequence from the dictionary, as exemplified by the System Prompt in Listing~\ref{prompt}.

\begin{lstlisting}[caption=System Prompt Template,label={prompt}]
You are a PRECISE text decoder. 
Replace ALL <M###> tokens with EXACT 
dictionary values. 

Dictionary: {dictionary}

RULES:
1. Find EVERY <M###> token and replace with its EXACT dictionary value
2. Copy ALL other text EXACTLY as written
3. NEVER modify any content except <M###> tokens
4. Output EVERY character from input

REPLACE ALL TOKENS. 
PRESERVE ALL OTHER TEXT.

\end{lstlisting}


Successful decompression demonstrates that the LLM has learned the dictionary and can apply it correctly, serving as a proxy for the LLM's ability to understand and analyze compressed data.
We test both Claude 3.7 Sonnet and Nova Premier models on this decompression task. Batch sizes are determined by the decompressed output size to prevent output truncation: datasets exceeding the LLM output context window limits (64k tokens for Claude 3.7 Sonnet, 32k tokens for Nova Premier) are processed in multiple batches.

\textbf{Evaluation Metrics}

We employ multiple metrics to assess decompression accuracy: \textit{Exact Match} (binary metric indicating perfect reconstruction, primary metric), \textit{Levenshtein Similarity} (character-level edit distance), \textit{Hamming Similarity} (position-wise character matching), \textit{ROUGE Scores} (n-gram recall, ROUGE-1 and ROUGE-L variants), \textit{BLEU Score} (n-gram precision), and \textit{String Presence} (verification that expected substrings appear in output).

For the LogHub dataset, the template-based compression approach compresses each log independently, resulting in approximately one meta-token per log. Although logs are sent to the LLM together in batches (constrained by output context window limits), the per-log compression enables direct one-to-one mapping between original and decompressed logs. Metrics are computed per log (2000 measurements per dataset), and we report mean $\pm$ standard error of the mean (SEM) to quantify the precision of our estimate of the true mean performance. 

\subsubsection{Algorithm-Based Decompression}\label{sec:sec:sec:algo-decompression}

\textbf{Experimental Design}

We evaluate our compression pipeline using Algorithm~\ref{alg:hierarchical_compression} on 2000 logs per dataset with Claude 3.7 Sonnet. This experiment tests end-to-end performance: compression using our hierarchical algorithm, dictionary generation, and LLM analysis of compressed data. We measure compression ratios for each dataset across $L_{max}$ values from 3 to 10.

\textbf{Evaluation Metrics}


Unlike template-based compression, the hierarchical algorithm identifies patterns across the entire batch rather than compressing each log independently. This cross-log compression is what enables high compression ratios, but it also prevents straightforward one-to-one mapping between original and decompressed logs, requiring batch-level evaluation. For some datasets, a single batch contained all 2000 logs, meaning even one decompression mistake yields an exact match score of zero. Consequently, exact match is overly strict for this evaluation, making similarity-based metrics more appropriate.

We focus on: \textit{Compression Ratio} (token reduction achieved), \textit{Levenshtein Similarity} (character-level accuracy), \textit{ROUGE Scores} (content recall), and \textit{BLEU Scores} (content precision). Each metric is computed once per (dataset, $L_{max}$) configuration on the entire batch of 2000 logs. Since we evaluate 8 different $L_{max}$ values (3--10) per dataset, we report mean $\pm$ standard deviation (std) to characterize the variability of performance across compression levels.

\section{Results}\label{sec:results}

\subsection{Compression Efficiency}
\label{sec:sec:compressioneff}

\begin{figure*}[h!]
    \centering
    \includegraphics[width=2.0\columnwidth]{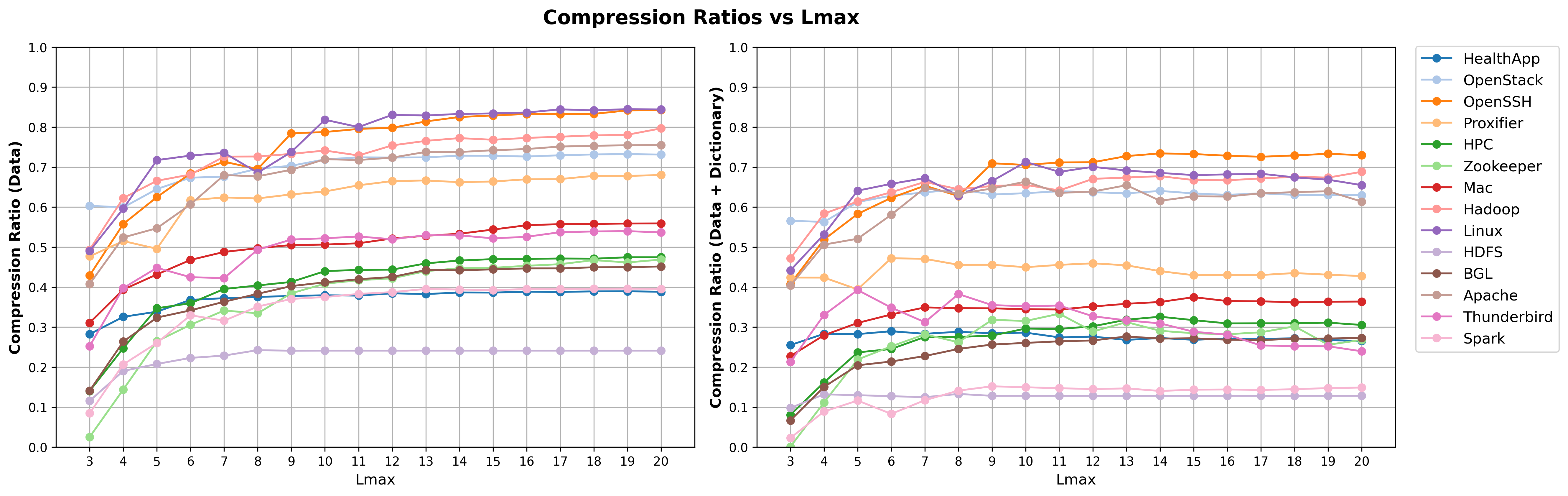}
    \caption{Compression ratios vs. $L_{max}$ without dictionary overhead (right) and with dictionary overhead (left). Most datasets show plateau behavior, while some exhibit optimal $L_{max}$ values beyond which dictionary overhead reduces compression efficiency.}
    \label{fig:image1}
\end{figure*}

The amount of compression obtained for each dataset depends on its intrinsic repetition patterns. Figure~\ref{fig:image1} illustrates compression performance of each dataset across different values of $L_{max}$, with the right side showing compression ratios without dictionary overhead and the left side accounting for dictionary size.

Looking at the compressed data in isolation (without accounting for the size of the meta-token dictionary), we observe that as $L_{max}$ increases, the compression ratio also increases until it plateaus. This behavior reflects the algorithm's ability to capture longer repetitive patterns, with diminishing returns as pattern lengths exceed the natural repetition characteristics of the data.

When accounting for the meta-token dictionary that must be included in the system prompt, most datasets maintain the plateau behavior, with compression ratios reduced by $0.1-0.2$ due to dictionary size. However, certain datasets such as Proxifier, Thunderbird and Zookeeper exhibit a distinct pattern: there exists an optimal $L_{max}$ that maximizes compression ratio, beyond which performance declines. This occurs because the token cost of encoding longer, rarer subsequences in the dictionary begins to offset the compression gains from replacing those subsequences in the data. The decline sometimes appears linear, sometimes with local maxima, reflecting the complex interplay between pattern frequency, length, and the token savings condition.

Figure~\ref{fig:image1} also shows that average compression performance varies across datasets. Three distinct clusters emerge: \textit{low compression} (0.1-0.2) includes HDFS and Spark, both large-scale distributed systems; \textit{medium compression} (0.2-0.5) encompasses BGL, HealthApp, HPC, Mac, Proxifier, Thunderbird, and Zookeeper, representing mixed system types; and \textit{high compression} (0.6-0.8) includes Apache, Hadoop, Linux, OpenSSH, and OpenStack, predominantly server applications and operating systems. This clustering suggests that compression effectiveness correlates with log structure regularity, where server applications and system logs seem to exhibit more predictable patterns than distributed computing frameworks.

\subsection{Template-Based Decompression}

\subsubsection{LogHub 2k Dataset Results}

\begin{figure*}[h!]
    \centering
    \includegraphics[width=2.0\columnwidth]{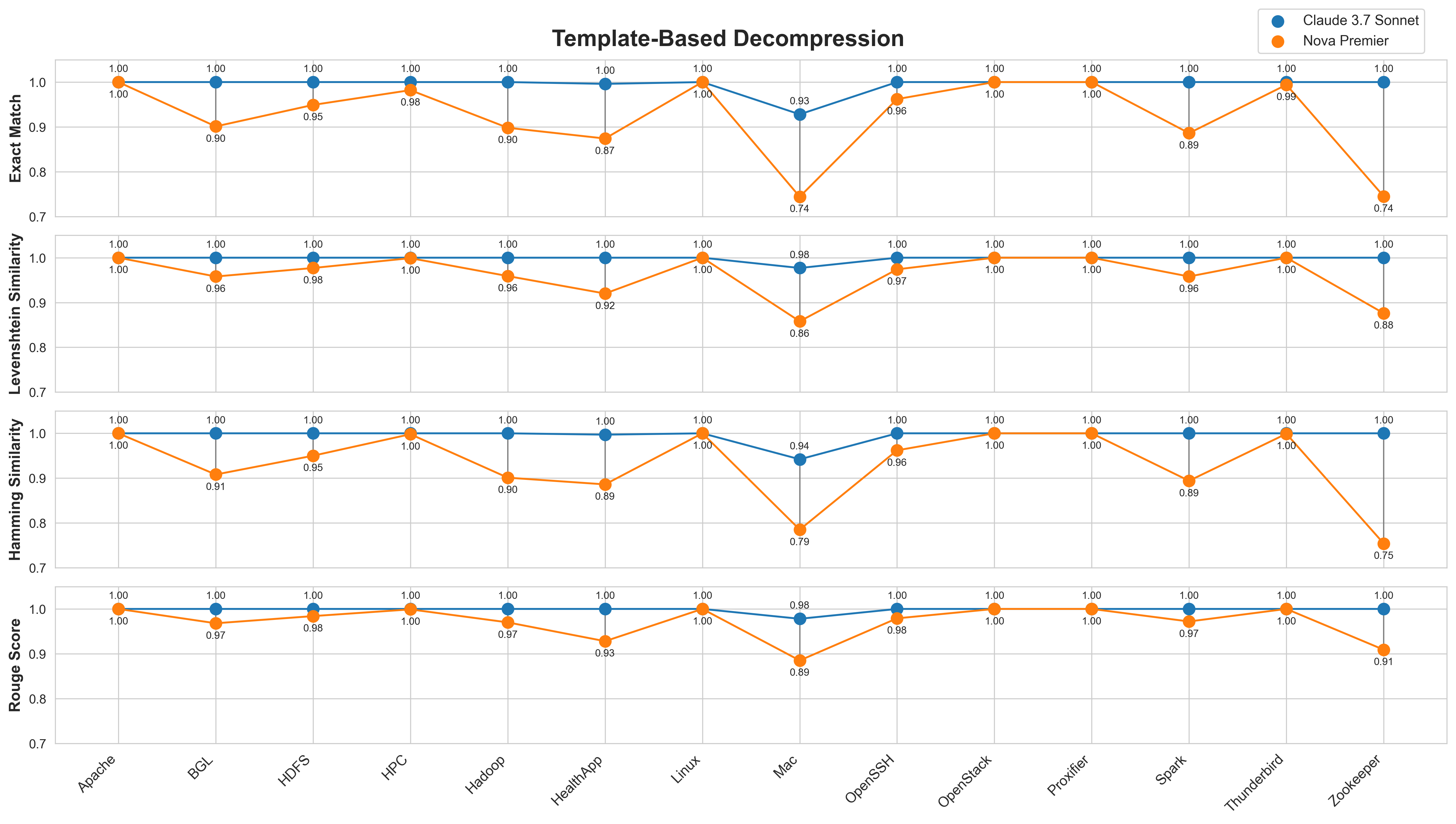}
    \caption{Template-based decompression results comparing Claude 3.7 Sonnet and Nova Premier across LogHub 2k datasets. Each point represents the average metric score computed over 2,000 logs per dataset.}
    \label{fig:template-decompression}
\end{figure*}

\begin{table*}[h!]
\centering
\caption{LogHub Full Dataset Template-Based Results (Claude 3.7 Sonnet)}
\label{tab:loghubfull}
\begin{tabular}{lcccccc}
\toprule
Dataset & Entries & Templates & Exact Match & Levenshtein & Hamming & Rouge  \\
\midrule
Apache & 51,978 & 29 & 0.999 & 0.999 & 0.999 & 0.999 \\
Mac & 100,314 & 626 & 0.969 & 0.987 & 0.975 & 0.988\\
\bottomrule
\end{tabular}
\end{table*}

Figure~\ref{fig:template-decompression} presents the template-based decompression results for Claude 3.7 Sonnet and Nova Premier across all 14 datasets. Both models demonstrate strong decompression capabilities, though with some performance differences.

\textbf{Claude 3.7 Sonnet} achieves exceptional performance, with exact match scores of 1.000 across 12 of 14 datasets (mean $0.994 \pm 0.018$ across datasets), see Appendix~\ref{appendix:template_metrics} for detailed per-dataset breakdown. The Mac dataset represents the lower bound at 0.928 exact match. Even for Mac, Levenshtein similarity remains high at 0.977, indicating that errors are typically minor character-level discrepancies rather than fundamental failures.

\textbf{Nova Premier} achieves exact match scores of 1.000 across 4 of 14 datasets: Apache, Linux, OpenStack, and Proxifier. The remaining datasets exhibit lower exact match rates, with Mac again representing the lower bound across both models (mean $0.924 \pm 0.086$ across all datasets). Detailed per-dataset metrics with SEM are provided in Appendix~\ref{appendix:template_metrics}.
Although Nova Premier exhibits lower exact match rates than Claude on several datasets, it still maintains good similarity scores. For the lowest-performing dataset (Mac), Levenshtein similarity remains at 0.858 and ROUGE score at 0.885, indicating substantial content preservation.

\subsubsection{Full Dataset Results}

Based on the 2k dataset results, we selected two contrasting systems for full-scale evaluation using Claude 3.7 Sonnet (the better-performing model): Apache, representing a well-performing dataset, and Mac, the lowest-performing dataset. Their substantial differences in size (52k vs.\ 100k entries) and template count (29 vs.\ 626 templates) provide a test of scalability across both dataset volume and dictionary complexity.

Table~\ref{tab:loghubfull} shows that performance remains high on both full datasets. Apache maintains near-perfect accuracy (0.999 exact match), while Mac achieves 0.969 exact match, a notable improvement over the 0.928 observed on the 2k subset. In addition, the 21$\times$ difference in dictionary size (29 vs.\ 626 templates) does not degrade performance, demonstrating that LLMs can effectively learn and apply substantially larger compression dictionaries without compromising the output.

\subsection{Algorithmic Compression Evaluation}

The template-based experiments in the previous section established that LLMs can learn dictionaries in-context and apply them accurately. We now evaluate whether this capability extends to our hierarchical compression algorithm, which applies substantially more compression by identifying patterns across the entire batch rather than individual logs.

\begin{figure*}[h!]
    \centering
    \includegraphics[width=2.0\columnwidth]{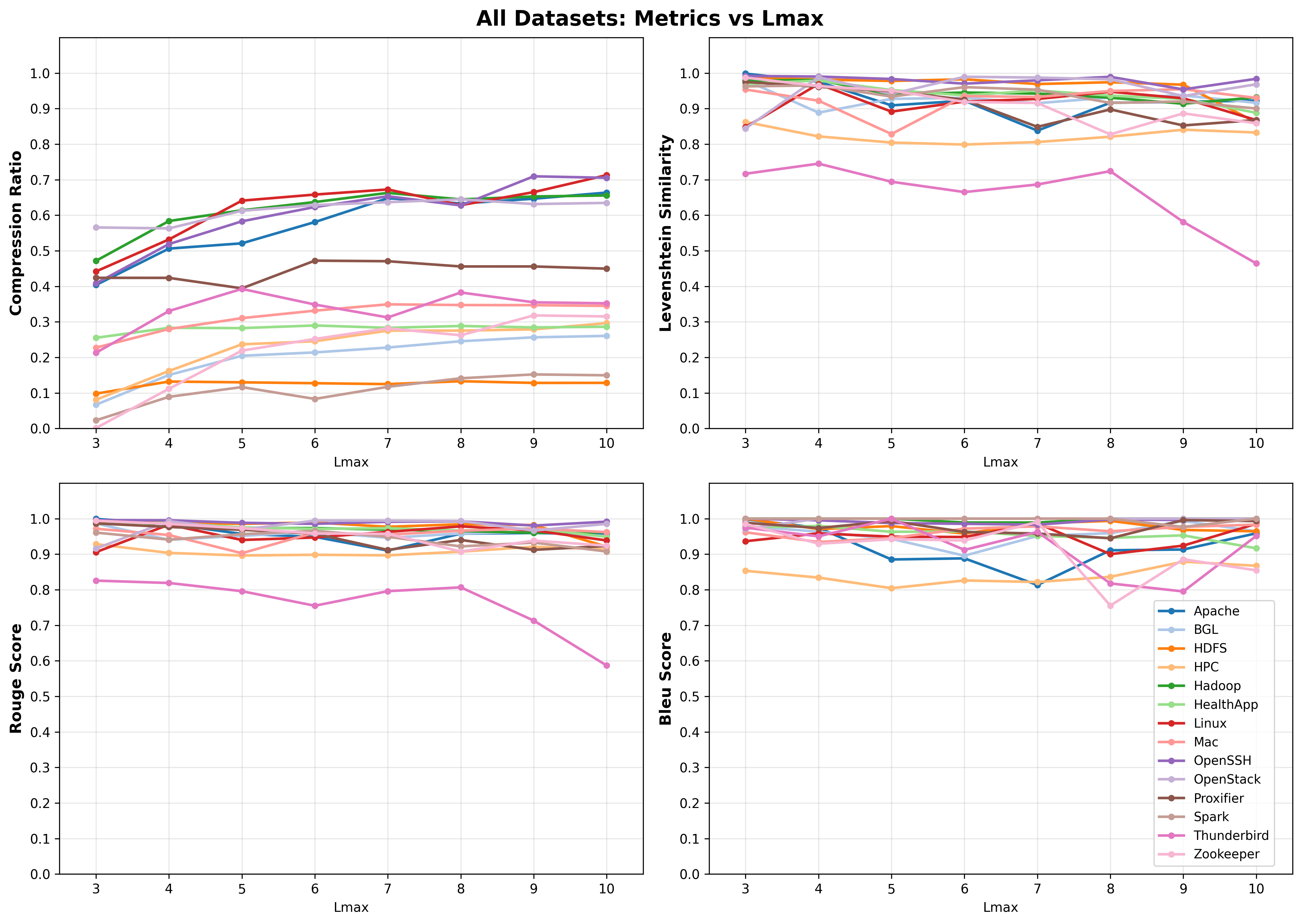}
    \caption{Compression ratios and decompression quality as a function of $L_{max}$. Top-left: compression ratios by dataset. Remaining panels: Levenshtein, ROUGE, and BLEU scores. Most datasets maintain scores above 0.9 regardless of compression level; outliers (HPC, Thunderbird) show consistently lower scores across all $L_{max}$ values, indicating dataset-specific rather than compression-related effects.}
\label{fig:similarity_scores}
\end{figure*}

Figure~\ref{fig:similarity_scores} shows compression ratios and decompression quality across all datasets and $L_{max}$ values 3--10. The top-left panel displays compression ratios (replicating Figure~\ref{fig:image1} for reference), while the remaining panels show similarity metrics. Datasets spanning the full range of compression ratios---from low (HDFS, Spark at 0.1--0.2) to high (Apache, OpenSSH at 0.6--0.8)---all maintain similarity scores above 0.90 on average. The two outliers, HPC and Thunderbird, exhibit lower scores despite only moderate compression ratios, suggesting that factors other than compression intensity drive reconstruction difficulty. Per-dataset breakdowns are provided in Appendix~\ref{appendix:detailed_metrics}.

\begin{table*}[h!]
\centering
\caption{Decompression quality for algorithmic compression across $L_{max}$ values 3--10. Metrics reported as mean $\pm$ std across compression levels. Most datasets (12/14) achieve Levenshtein similarity above 0.90; outliers (HPC, Thunderbird) reflect dataset-specific characteristics discussed in Section~\ref{sec:discussion}.}
\label{tab:algorithmic_per_dataset}
\begin{tabular}{lccc}
\toprule
\textbf{Dataset} & \textbf{Levenshtein} & \textbf{ROUGE} & \textbf{BLEU} \\
\midrule
Apache & $0.925 \pm 0.048$ & $0.959 \pm 0.026$ & $0.918 \pm 0.059$ \\
BGL & $0.928 \pm 0.026$ & $0.957 \pm 0.014$ & $0.956 \pm 0.029$ \\
Hadoop & $0.945 \pm 0.023$ & $0.972 \pm 0.010$ & $0.997 \pm 0.005$ \\
HDFS & $0.963 \pm 0.042$ & $0.978 \pm 0.023$ & $0.978 \pm 0.015$ \\
HealthApp & $0.943 \pm 0.028$ & $0.973 \pm 0.012$ & $0.956 \pm 0.020$ \\
HPC & $0.823 \pm 0.021$ & $0.908 \pm 0.012$ & $0.840 \pm 0.025$ \\
Linux & $0.912 \pm 0.041$ & $0.953 \pm 0.025$ & $0.949 \pm 0.029$ \\
Mac & $0.926 \pm 0.041$ & $0.956 \pm 0.022$ & $0.964 \pm 0.017$ \\
OpenSSH & $0.980 \pm 0.013$ & $0.990 \pm 0.005$ & $0.993 \pm 0.007$ \\
OpenStack & $0.955 \pm 0.049$ & $0.977 \pm 0.027$ & $0.996 \pm 0.010$ \\
Proxifier & $0.910 \pm 0.050$ & $0.947 \pm 0.030$ & $0.976 \pm 0.019$ \\
Spark & $0.939 \pm 0.025$ & $0.942 \pm 0.020$ & $0.997 \pm 0.008$ \\
Thunderbird & $0.660 \pm 0.093$ & $0.762 \pm 0.080$ & $0.920 \pm 0.075$ \\
Zookeeper & $0.914 \pm 0.054$ & $0.955 \pm 0.030$ & $0.910 \pm 0.077$ \\
\midrule
\textbf{Aggregate} & $\mathbf{0.909 \pm 0.088}$ & $\mathbf{0.945 \pm 0.061}$ & $\mathbf{0.954 \pm 0.055}$ \\
\bottomrule
\end{tabular}
\end{table*}

\begin{figure*}[h!]
    \centering
    \includegraphics[width=2.\columnwidth]{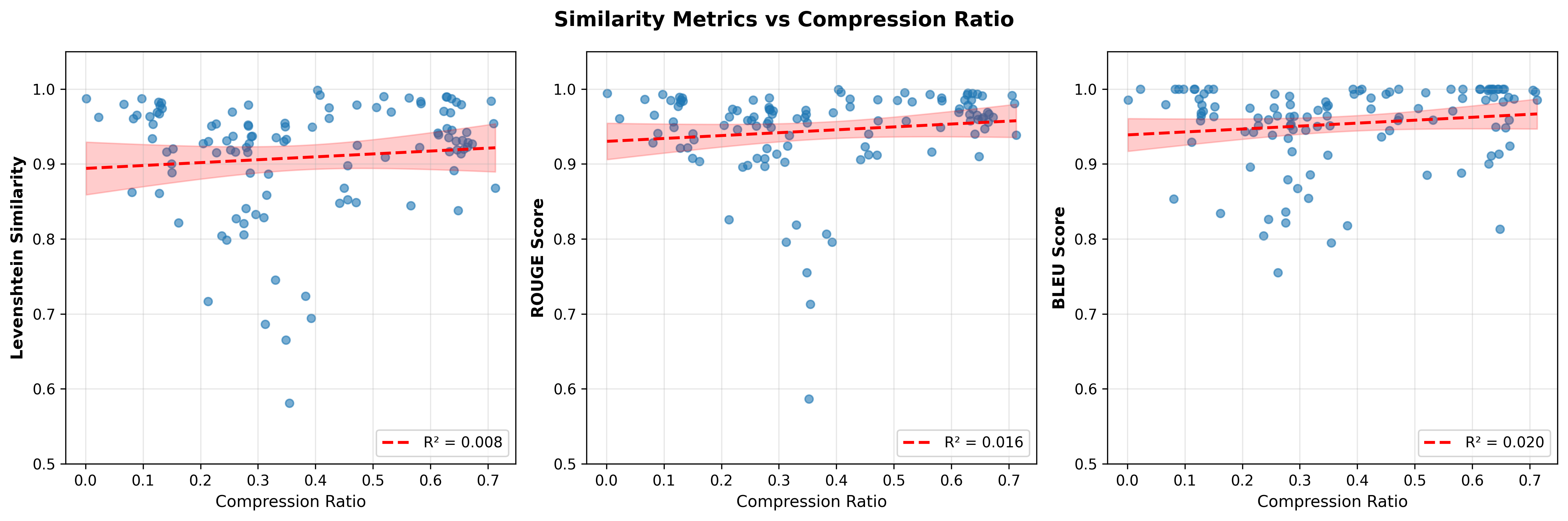}
    \caption{Relationship between compression ratio and decompression quality. Each point represents one (dataset, $L_{max}$) configuration. Dashed lines show linear regression fits; shaded regions indicate 95\% confidence intervals. R$^2$ values below 0.02 for all metrics demonstrate that compression intensity does not predict reconstruction quality---performance variation is driven by dataset characteristics rather than compression level.}
    \label{fig:compression_correlation}
\end{figure*}

Table~\ref{tab:algorithmic_per_dataset} summarizes these results, averaging metrics over $L_{max}$ values 3--10. The majority of datasets achieve excellent reconstruction: 12 of 14 exceed 0.90 Levenshtein similarity, with aggregate scores of $0.909 \pm 0.088$ (Levenshtein), $0.945 \pm 0.061$ (ROUGE), and $0.954 \pm 0.055$ (BLEU).

To formally test whether compression ratio affects decompression quality, we performed linear regression analysis across all (dataset, $L_{max}$) configurations. Figure~\ref{fig:compression_correlation} plots each similarity metric against compression ratio, with regression lines and 95\% confidence intervals. The analysis yields R$^2$ values of 0.008 (Levenshtein), 0.016 (ROUGE), and 0.020 (BLEU), indicating that compression ratio explains less than 2\% of variance in any metric. The near-horizontal regression lines confirm that decompression quality remains consistently high regardless of compression intensity.

These results lead to the finding that the LLM's ability to decompress data depends on intrinsic dataset properties rather than the degree of compression applied. High compression ratios do not degrade performance, and the outlier datasets (HPC, Thunderbird) perform poorly regardless of $L_{max}$. We analyze the structural characteristics that explain these outliers in Section~\ref{sec:discussion}.

\section{Discussion}
\label{sec:discussion}

\subsection{Mechanism of In-Context Dictionary Learning}

The high decompression accuracy observed across experiments indicates that LLMs treat dictionary-based compression as a deterministic symbol substitution task. Unlike lossy compression methods that require models to infer missing information, dictionary-encoding preserves semantic completeness since each meta-token maps unambiguously to its original content. This explains why performance remains stable as dictionary size increases from 29 to 626 entries and dataset size scales from 2,000 to over 100,000 logs: the underlying task complexity does not change with scale.

The absence of correlation between compression ratio and reconstruction quality (R$^2 < 0.02$) further supports this interpretation. Whether a document contains few or many meta-tokens, each individual substitution remains equally interpretable. This independence has practical significance: compression can be maximized without quality trade-offs, with the limiting factor being data characteristics rather than compression intensity.

\subsection{Role of Semantic Structure in Reconstruction}

Performance variation across datasets reflects differences in data structure rather than compressibility. Of 14 datasets evaluated, 12 achieve Levenshtein similarity above 0.90. The two outliers---HPC (0.82) and Thunderbird (0.66)---share a common characteristic: dense alphanumeric sequences without surrounding natural language context.

Thunderbird logs contain standalone node identifier lists (e.g., ``an16 an17 an18...an125'') where multiple dictionary entries differ only in the specific numeric range (``an16--an25'', ``an26--an35'', etc.). These near-identical sequences provide no distinguishing context for accurate reconstruction. HPC logs similarly include repetitive status matrices and numeric sequences with minimal descriptive text. These patterns compress effectively but provide no contextual cues during reconstruction. In contrast, datasets like OpenStack and Hadoop embed complex identifiers (UUIDs, task IDs) within descriptive phrases such as ``Progress of TaskAttempt attempt\_... is :'' that guide accurate reconstruction.

This suggests that compression effectiveness depends on co-occurrence of repetitive patterns with natural language structure, a condition satisfied by the majority of enterprise log data.

\subsection{Practical Implications}

These findings have implications for LLM deployment:

\textbf{Cost reduction}: Compression ratios of 60\%--80\% translate directly to proportional reductions in API costs, enabling analyses of large-scale datasets that would otherwise be cost-prohibitive.

\textbf{Context utilization}: Reduced token consumption allows more data to fit within fixed context windows, enabling comprehensive analysis without truncation or sampling.

\textbf{Deployment simplicity}: The approach requires no model fine-tuning or specialized infrastructure, allowing immediate deployment with existing LLM integrations.

\textbf{Generalization}: Beyond system logs, the method applies to any domain with repetitive textual patterns, e.g., process mining, software engineering artifacts, templated documents, and structured reports. Domains with high structural regularity will see the greatest benefit.

\section{Limitations and Future Work}\label{sec:limitations}

Our evaluation relies on decompression as a proxy for the LLM's ability to understand compressed data in context. While high decompression accuracy suggests the model has internalized the dictionary mappings, we cannot directly observe the internal representations of API-based LLMs to verify that compression introduces no analytical loss. The assumption that accurate decompression implies accurate analysis is reasonable but remains empirically validated rather than proven.

Importantly, decompression is a stricter test than understanding since it requires perfect reconstruction of the original text. An LLM may fully understand the compressed content yet produce minor errors when generating the decompressed output given that reconstruction is itself an additional generative task with its own error modes. Consequently, our similarity metrics likely underestimate the LLM's true comprehension of compressed data, and analytical performance may exceed what decompression accuracy alone suggests.

Future work should investigate the relationship between compression patterns and LLM architectural characteristics, potentially revealing which model families or attention mechanisms are best suited for dictionary-based compression. In addition, extending the evaluation to other downstream analytical tasks would strengthen the connection between decompression fidelity and practical utility.

\section{Conclusion}
\label{sec:conclusion}

We present a training-free approach to lossless prompt compression that leverages LLM in-context learning capabilities. Our experiments demonstrate that LLMs can learn compression dictionaries provided in the system prompt and accurately analyze compressed data, achieving exact match rates exceeding 0.99 for template-based compression and Levenshtein similarity scores above 0.91 on average for algorithmic compression across the LogHub 2.0 dataset. Compression ratios of 60\%--80\% do not impair analytical capability. Linear regression analysis shows compression ratio explains less than 2\% of variance in similarity metrics, with dataset characteristics rather than compression level determining performance.

These findings address a gap in LLM-based analysis of repetitive data. Existing compression methods either require costly fine-tuning that becomes impractical as datasets evolve, or sacrifice analytical fidelity through lossy compression. By demonstrating that LLMs can learn compression dictionaries through in-context learning alone, we enable lossless input compression without model modifications. The approach requires no specialized infrastructure and works with standard API-based LLMs.

For organizations processing large volumes of repetitive data, this translates directly to cost savings: 60\%--80\% compression ratios mean proportional reduction in token-based API costs, enabling comprehensive analyses that would otherwise be cost-prohibitive. As LLM adoption grows across enterprise applications, dictionary-based compression offers a practical path to scaling these capabilities economically.

\section*{Acknowledgements}

We thank Dave George, Viven Gupta, and Stephen Lau for their support in the writing of this manuscript.
We also thank our colleagues from Amazon Fintelligence, Aditya Paul, Rohan Singh, Zishan Muzeeb, Aitzaz Ahmad, Karan Aggarwal, Yong Xie, Wei Guo, Venkatesh Kaulgud, Gaini Kussainova, and Xiuhong Du for multiple useful discussions. 

\nocite{langley00}

\bibliography{example_paper}

@inproceedings{langley00,
 author    = {P. Langley},
 title     = {Crafting Papers on Machine Learning},
 year      = {2000},
 pages     = {1207--1216},
 editor    = {Pat Langley},
 booktitle     = {Proceedings of the 17th International Conference
              on Machine Learning (ICML 2000)},
 address   = {Stanford, CA},
 publisher = {Morgan Kaufmann}
}

@article{harvill2025lossless,
  title={Lossless Token Sequence Compression via Meta-Tokens},
  author={Harvill, Adam and others},
  journal={arXiv preprint arXiv:2506.00307},
  year={2025},
  url={https://arxiv.org/abs/2506.00307}
}

@article{zhang2024incontext,
  title={In-Context Former: Lightning-fast Context Compression for LLMs},
  author={Zhang, Tianjun and others},
  journal={arXiv preprint arXiv:2403.18119},
  year={2024},
  url={https://arxiv.org/abs/2403.18119}
}

@article{hou2024instruction,
  title={Instruction-Aware Contextual Compression for Long Context Large Language Models},
  author={Hou, Le and others},
  journal={arXiv preprint arXiv:2402.03229},
  year={2024},
  url={https://arxiv.org/abs/2402.03229}
}

@article{xu2024xraglog,
  title={XRAGLog: Retrieval-Augmented Log Anomaly Detection with Large Language Models},
  author={Xu, Yuxuan and others},
  journal={arXiv preprint arXiv:2401.00889},
  year={2024},
  url={https://arxiv.org/abs/2401.00889}
}

@article{zhu2023loghub,
  title={Loghub: A large collection of system log datasets towards automated log analytics},
  author={Zhu, Xiaoqing and others},
  journal={EMSE},
  volume={28},
  pages={54},
  year={2023},
  doi={10.1186/s40411-023-00194-9},
  url={https://doi.org/10.1186/s40411-023-00194-9}
}

@article{li2023logshrink,
  title={Logshrink: Mining log invariants for lightweight compression},
  author={Li, Chendi and others},
  journal={IEEE Transactions on Software Engineering},
  year={2023},
  doi={10.1109/TSE.2023.3274956},
  url={https://doi.org/10.1109/TSE.2023.3274956}
}

@article{mao2024llmzip,
  title={LLMZip: Lossless compression of language model outputs via next-token prediction},
  author={Mao, Yutian and others},
  journal={arXiv preprint arXiv:2404.13529},
  year={2024},
  url={https://arxiv.org/abs/2404.13529}
}

@article{narasimhan2024alphazip,
  title={AlphaZip: Transformer-based compression of language model outputs},
  author={Narasimhan, Karthik and Chandrachoodan, N},
  journal={arXiv preprint arXiv:2404.02352},
  year={2024},
  url={https://arxiv.org/abs/2404.02352}
}

@article{jiang2024loghub2,
  title={A Large-scale Evaluation for Log Parsing Techniques: How Far are We?},
  author={Jiang, Zhihan and Liu, Jinyang and Huang, Junjie and Li, Yichen and Huo, Yintong and Gu, Jiazhen and Chen, Zhuangbin and Zhu, Jieming and Lyu, Michael R.},
  journal={ISSTA},
  year={2024},
  url={https://arxiv.org/abs/2308.10828}
}

@Book{Salomon:2004:DCC,
  title={Data Compression: The Complete Reference},
  author={Salomon, D.},
  isbn={9780387406978},
  lccn={2003065725},
  url={https://books.google.com/books?id=PT1fcX321I4C},
  year={2004},
  publisher={Springer New York}
}

@misc{jiang2023llmlingua,
      title={LLMLingua: Compressing Prompts for Accelerated Inference of Large Language Models}, 
      author={Huiqiang Jiang and Qianhui Wu and Chin-Yew Lin and Yuqing Yang and Lili Qiu},
      year={2023},
      eprint={2310.05736},
      archivePrefix={arXiv},
      primaryClass={cs.CL},
      url={https://arxiv.org/abs/2310.05736}, 
}

@misc{pan2024llmlingua2,
      title={LLMLingua-2: Data Distillation for Efficient and Faithful Task-Agnostic Prompt Compression}, 
      author={Zhuoshi Pan and Qianhui Wu and Huiqiang Jiang and Menglin Xia and Xufang Luo and Jue Zhang and Qingwei Lin and Victor Rühle and Yuqing Yang and Chin-Yew Lin and H. Vicky Zhao and Lili Qiu and Dongmei Zhang},
      year={2024},
      eprint={2403.12968},
      archivePrefix={arXiv},
      primaryClass={cs.CL},
      url={https://arxiv.org/abs/2403.12968}, 
}
\bibliographystyle{mlsys2025}

\appendix
\appendix

\begin{table*}[t]
\centering
\caption{Original Token Counts for LogHub-2k Datasets}
\label{tab:token_stats}
\begin{tabular}{lc}
\toprule
Dataset & Original Tokens \\
\midrule
HPC & 13,959 \\
Zookeeper & 27,878 \\
BGL & 32,951 \\
Apache & 36,449 \\
HealthApp & 37,533 \\
Spark & 39,604 \\
Thunderbird & 49,759 \\
Linux & 56,804 \\
OpenSSH & 57,213 \\
Hadoop & 66,275 \\
Mac & 74,719 \\
Proxifier & 77,061 \\
HDFS & 79,155 \\
OpenStack & 111,221 \\
\bottomrule
\end{tabular}
\end{table*}
\begin{table*}[h!]
\centering
\caption{Per-dataset template-based decompression metrics for Claude 3.7 Sonnet and Nova Premier on LogHub-2k (2000 logs per dataset). Per-dataset values reported as mean $\pm$ SEM; aggregate row reports mean $\pm$ std across datasets.}

\label{tab:template_combined}
\begin{tabular}{lcccccc}
\toprule
& \multicolumn{3}{c}{\textbf{Claude 3.7 Sonnet}} & \multicolumn{3}{c}{\textbf{Nova Premier}} \\
\cmidrule(lr){2-4} \cmidrule(lr){5-7}
\textbf{Dataset} & \textbf{Exact Match} & \textbf{Levenshtein} & \textbf{ROUGE} & \textbf{Exact Match} & \textbf{Levenshtein} & \textbf{ROUGE} \\
\midrule
Apache & $1.000$ & $1.000$ & $1.000$ & $1.000$ & $1.000$ & $1.000$ \\
BGL & $1.000$ & $1.000$ & $1.000$ & $0.901 \pm 0.007$ & $0.958 \pm 0.003$ & $0.968 \pm 0.003$ \\
Hadoop & $1.000$ & $1.000$ & $1.000$ & $0.898 \pm 0.007$ & $0.959 \pm 0.003$ & $0.970 \pm 0.003$ \\
HDFS & $1.000$ & $1.000$ & $1.000$ & $0.949 \pm 0.005$ & $0.977 \pm 0.002$ & $0.984 \pm 0.002$ \\
HealthApp & $0.996 \pm 0.001$ & $1.000$ & $1.000$ & $0.874 \pm 0.007$ & $0.920 \pm 0.005$ & $0.928 \pm 0.005$ \\
HPC & $1.000$ & $1.000$ & $1.000$ & $0.982 \pm 0.003$ & $0.999$ & $0.999$ \\
Linux & $1.000$ & $1.000$ & $1.000$ & $1.000$ & $1.000$ & $1.000$ \\
Mac & $0.928 \pm 0.006$ & $0.977 \pm 0.002$ & $0.978 \pm 0.003$ & $0.744 \pm 0.010$ & $0.858 \pm 0.006$ & $0.885 \pm 0.005$ \\
OpenSSH & $1.000$ & $1.000$ & $1.000$ & $0.962 \pm 0.004$ & $0.974 \pm 0.003$ & $0.979 \pm 0.003$ \\
OpenStack & $1.000$ & $1.000$ & $1.000$ & $1.000$ & $1.000$ & $1.000$ \\
Proxifier & $1.000$ & $1.000$ & $1.000$ & $1.000$ & $1.000$ & $1.000$ \\
Spark & $1.000$ & $1.000$ & $1.000$ & $0.886 \pm 0.007$ & $0.958 \pm 0.003$ & $0.972 \pm 0.002$ \\
Thunderbird & $1.000$ & $1.000$ & $1.000$ & $0.994 \pm 0.002$ & $1.000$ & $1.000$ \\
Zookeeper & $1.000$ & $1.000$ & $1.000$ & $0.745 \pm 0.010$ & $0.876 \pm 0.005$ & $0.909 \pm 0.004$ \\
\midrule
\textbf{Aggregate} & $\mathbf{0.994 \pm 0.018}$ & $\mathbf{0.998 \pm 0.006}$ & $\mathbf{0.998 \pm 0.006}$ & $\mathbf{0.924 \pm 0.086}$ & $\mathbf{0.963 \pm 0.046}$ & $\mathbf{0.971 \pm 0.036}$ \\
\bottomrule
\end{tabular}
\end{table*}

\section{Dataset Token Statistics}
\label{appendix:token_stats}

Table~\ref{tab:token_stats} presents the original token counts for each dataset in the LogHub-2k experiments, computed using the Claude tokenizer.

\section{Template-Based Decompression Detailed Metrics}
\label{appendix:template_metrics}

Table~\ref{tab:template_combined} presents per-dataset metrics for template-based decompression comparing Claude 3.7 Sonnet and Nova Premier. All metrics are reported as mean $\pm$ standard error of the mean (SEM), computed over n=2000 logs per dataset.

\section{Algorithmic Compression Detailed Metrics}
\label{appendix:detailed_metrics}

\begin{figure*}[t]
    \centering
    \includegraphics[width=1.8\columnwidth]{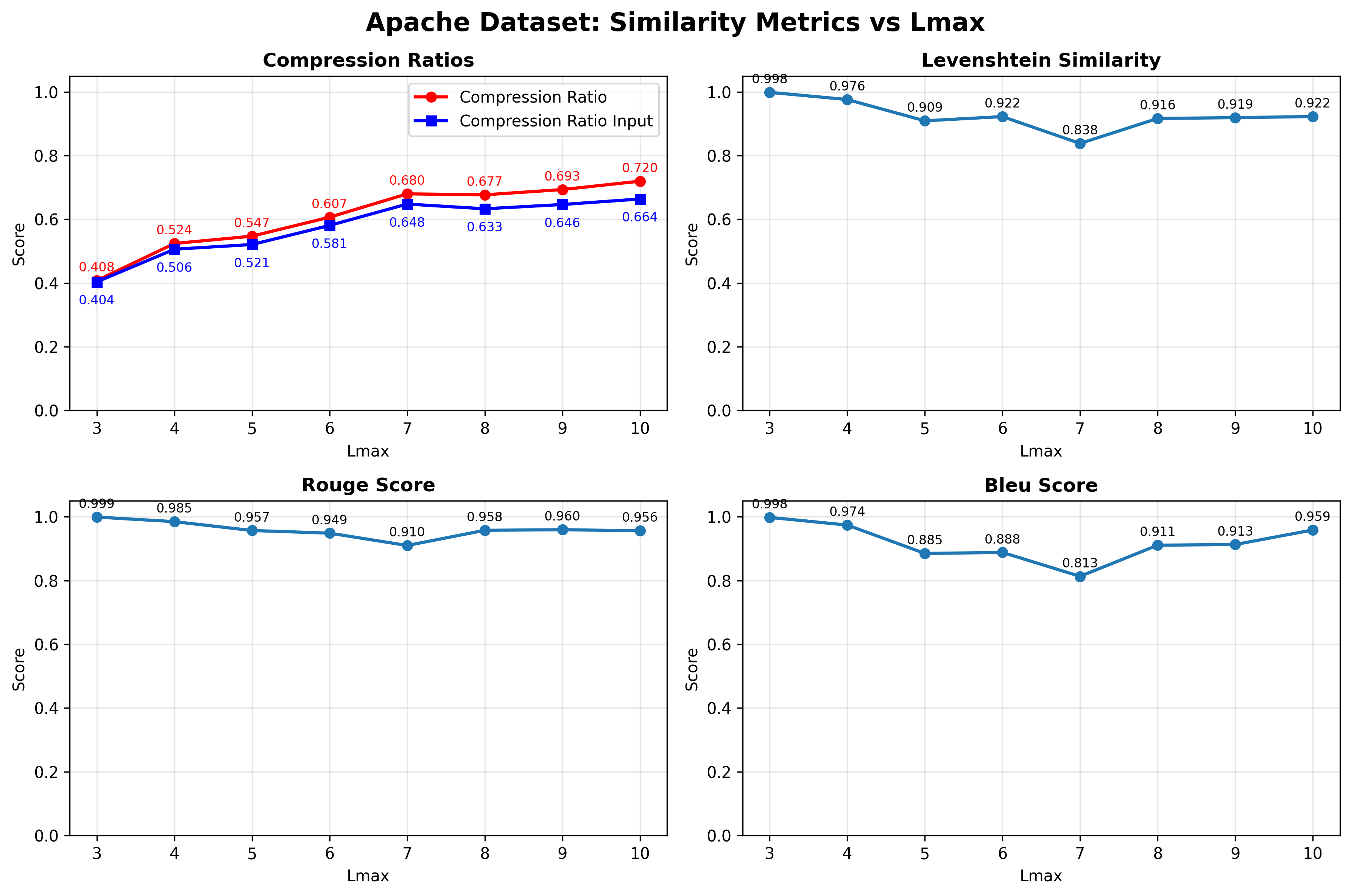}
    \caption{Levenshtein similarity, ROUGE, and BLEU scores for algorithmic compression for Apache dataset and $L_{max}$ values.}
    \label{fig:similarity_scores_apache}
\end{figure*}

\begin{figure*}[t]
    \centering
    \includegraphics[width=1.8\columnwidth]{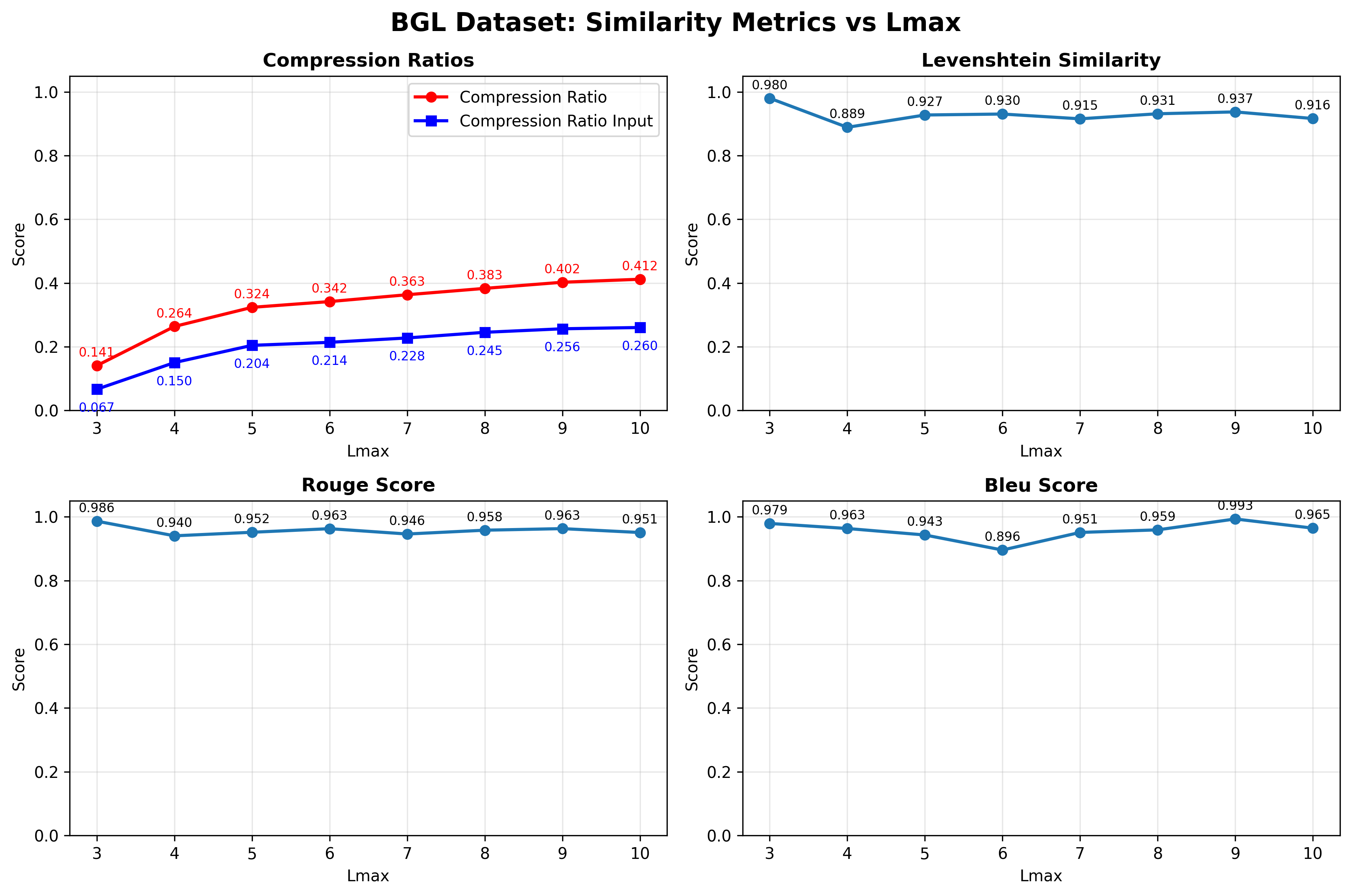}
    \caption{Levenshtein similarity, ROUGE, and BLEU scores for algorithmic compression for BGL dataset and $L_{max}$ values.}
    \label{fig:similarity_scores_bgl}
\end{figure*}

\begin{figure*}[t]
    \centering
    \includegraphics[width=1.8\columnwidth]{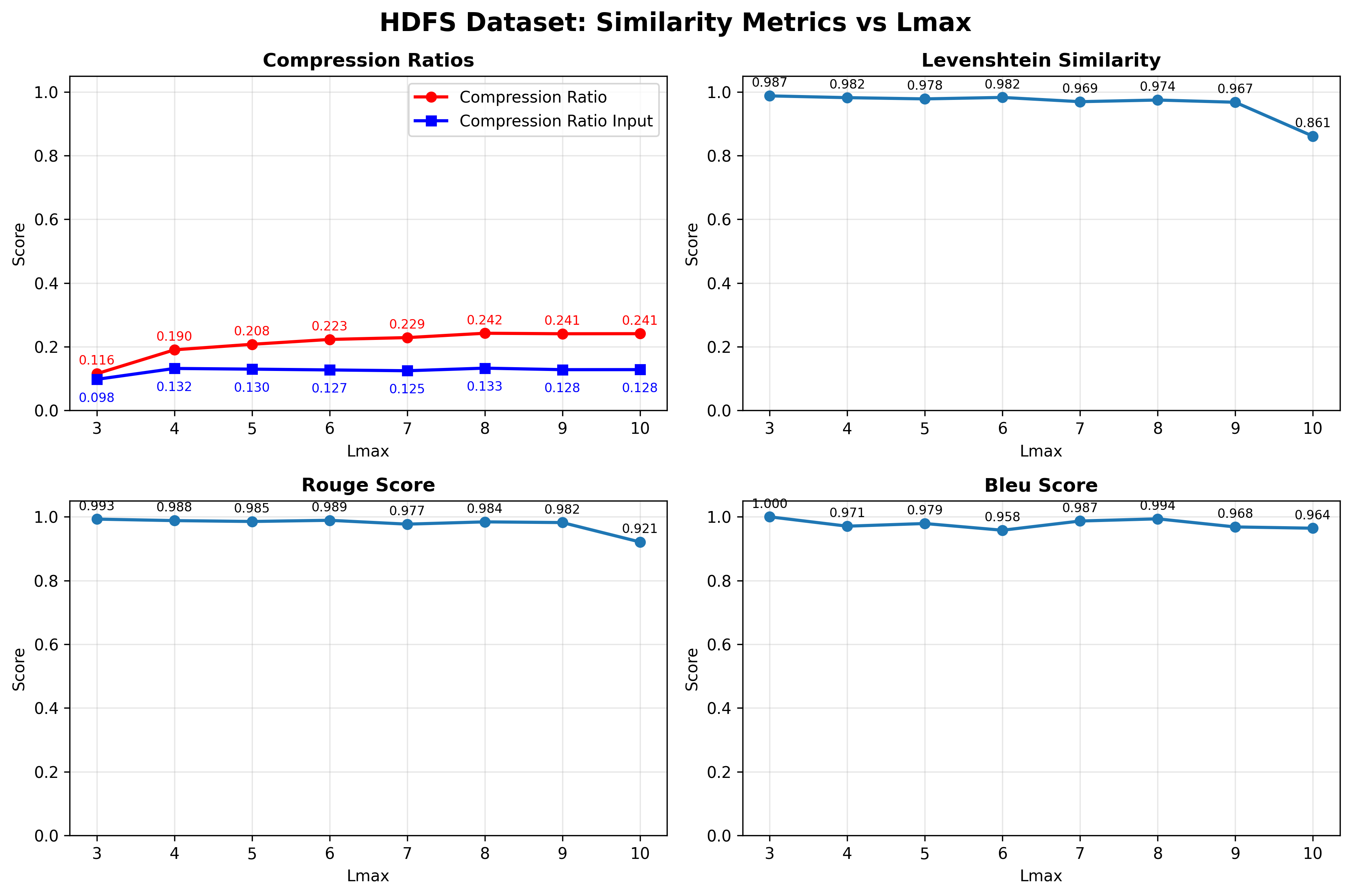}
    \caption{Levenshtein similarity, ROUGE, and BLEU scores for algorithmic compression for HDFS dataset and $L_{max}$ values.}
    \label{fig:similarity_scores_hdfs}
\end{figure*}

\begin{figure*}[t]
    \centering
    \includegraphics[width=1.8\columnwidth]{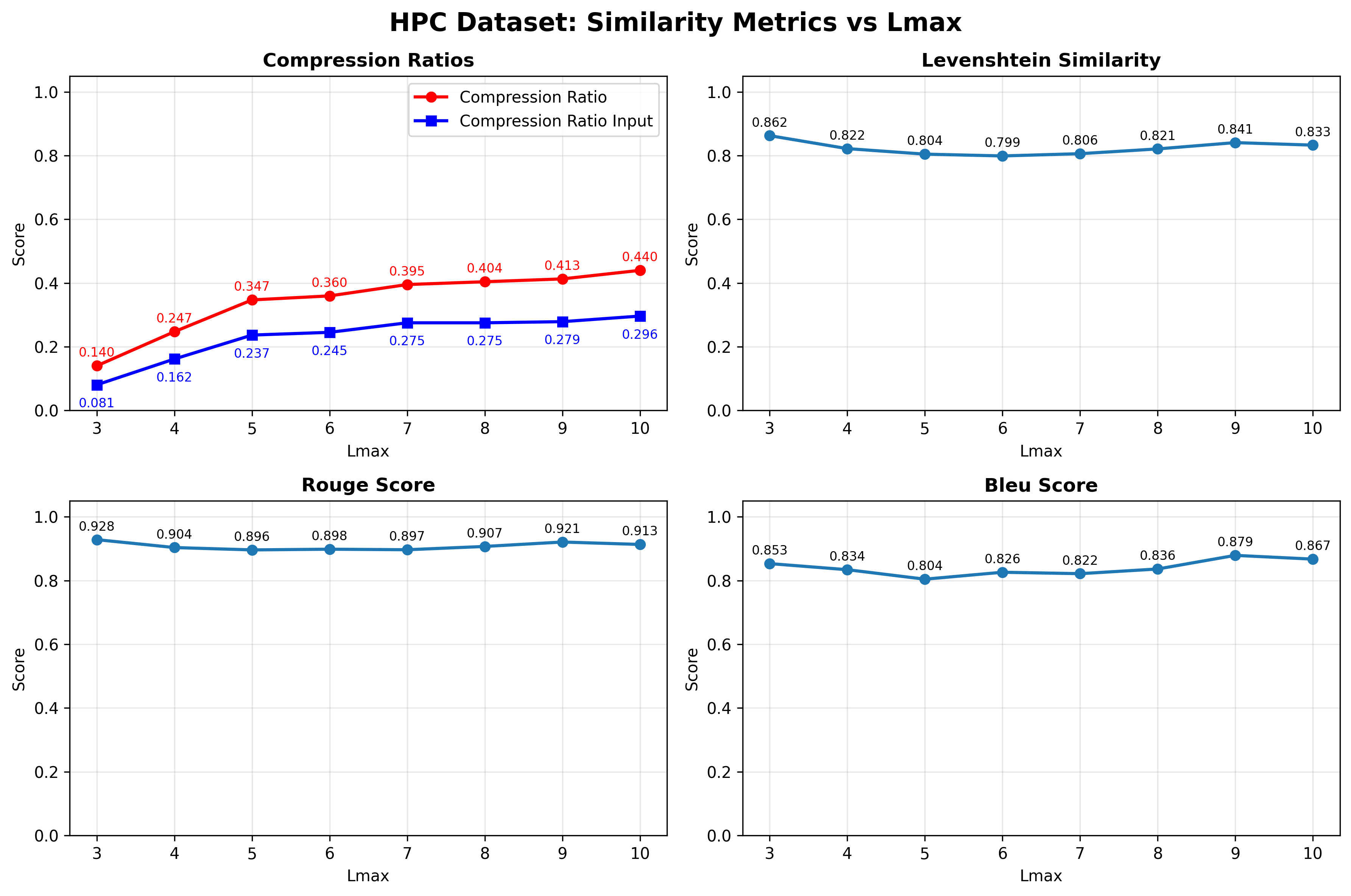}
    \caption{Levenshtein similarity, ROUGE, and BLEU scores for algorithmic compression for HPC dataset and $L_{max}$ values.}
    \label{fig:similarity_scores_hpc}
\end{figure*}

\begin{figure*}[t]
    \centering
    \includegraphics[width=1.8\columnwidth]{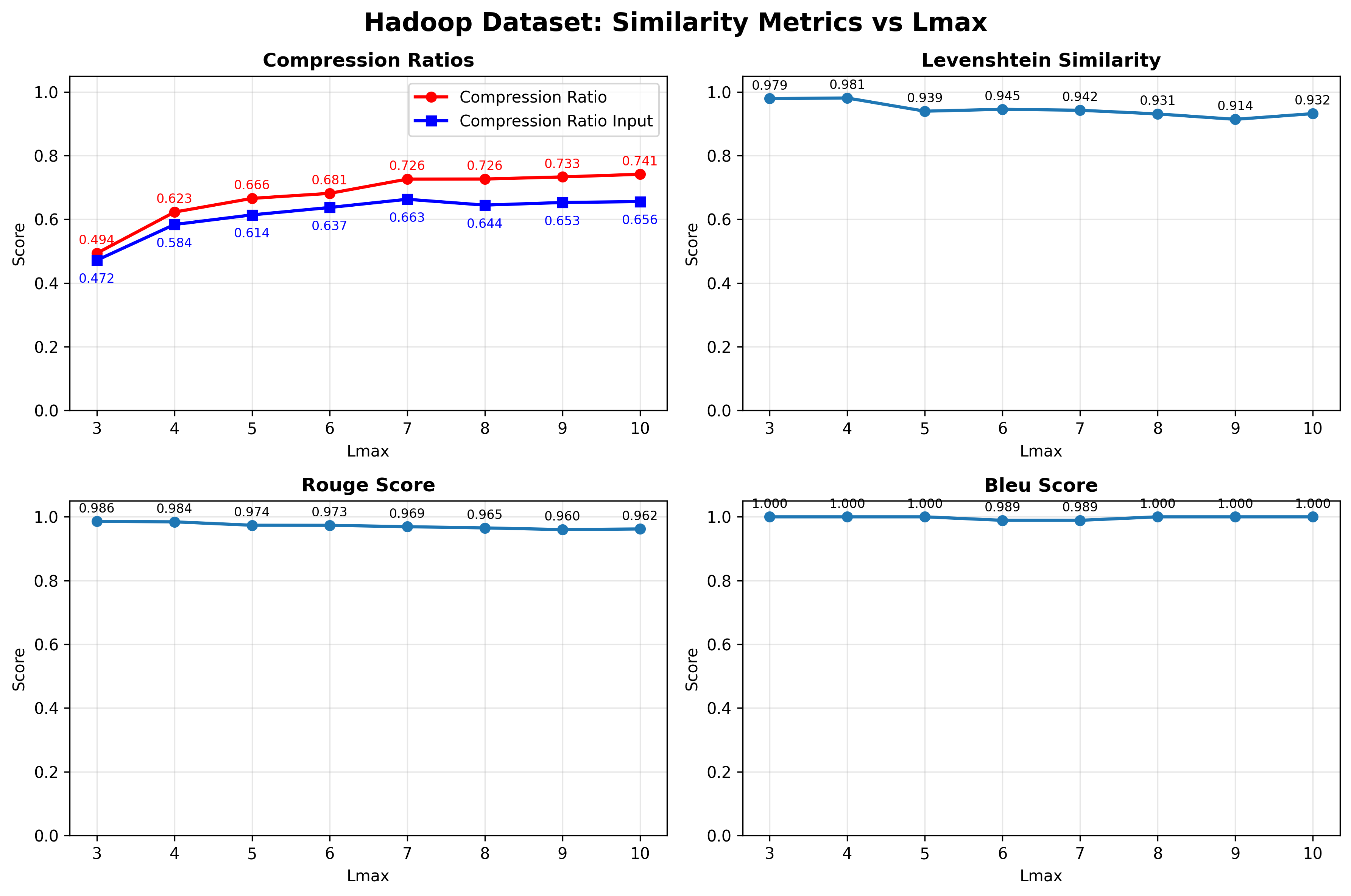}
    \caption{Levenshtein similarity, ROUGE, and BLEU scores for algorithmic compression for Hadoop dataset and $L_{max}$ values.}
    \label{fig:similarity_scores_hadoop}
\end{figure*}

\begin{figure*}[t]
    \centering
    \includegraphics[width=1.8\columnwidth]{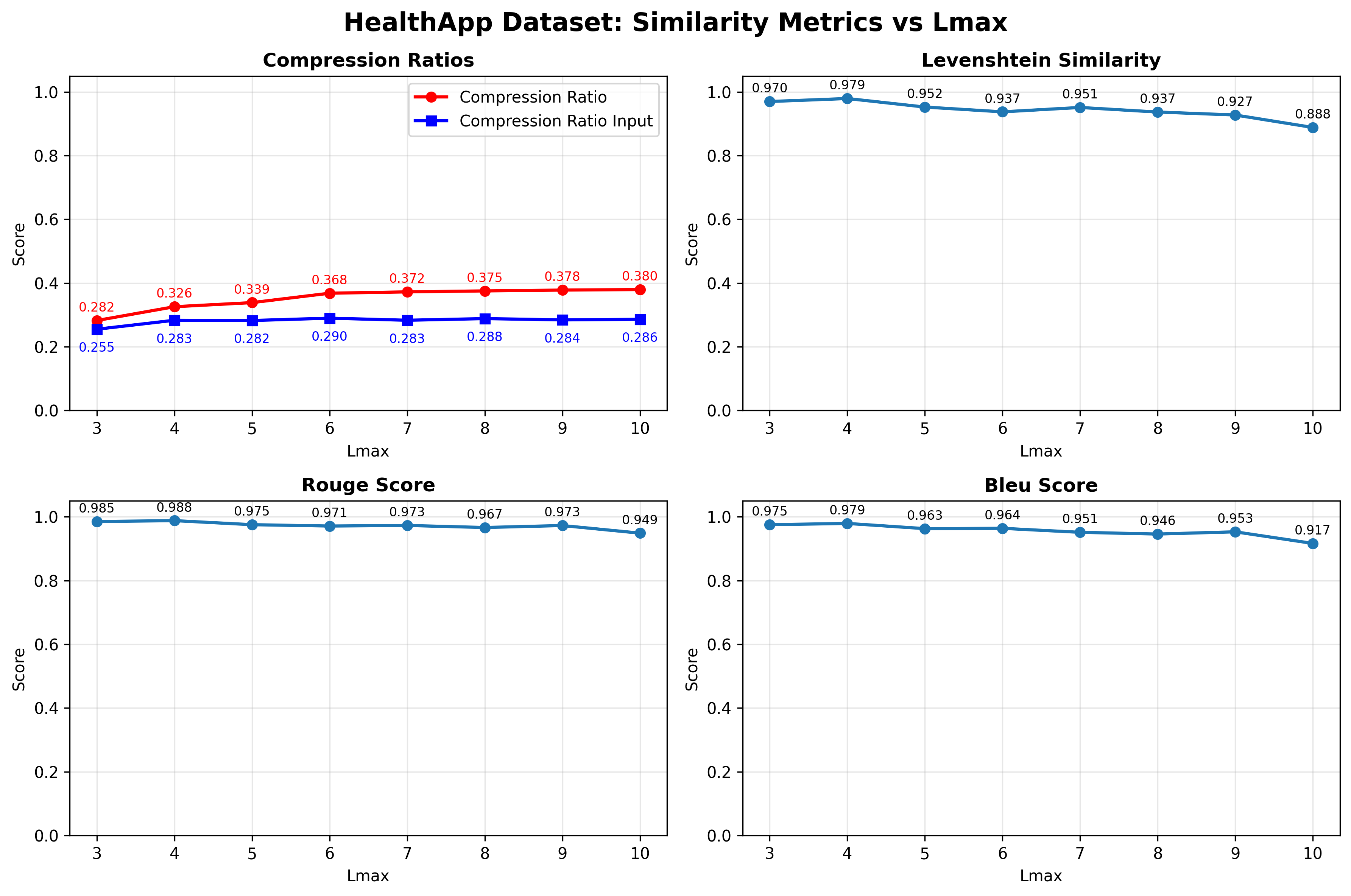}
    \caption{Levenshtein similarity, ROUGE, and BLEU scores for algorithmic compression for HealthApp dataset and $L_{max}$ values.}
    \label{fig:similarity_scores_healthapp}
\end{figure*}

\begin{figure*}[t]
    \centering
    \includegraphics[width=1.8\columnwidth]{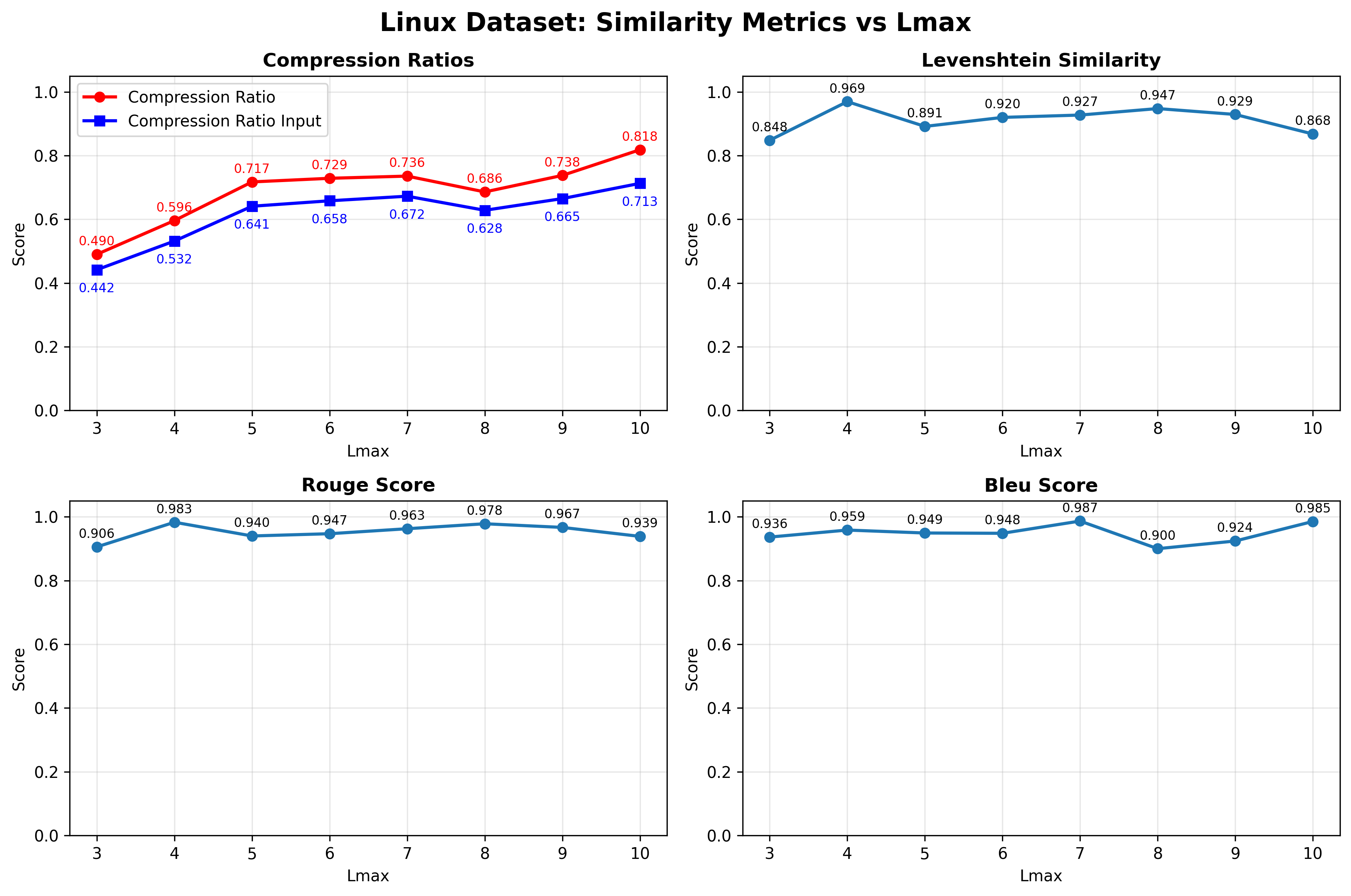}
    \caption{Levenshtein similarity, ROUGE, and BLEU scores for algorithmic compression for Linux dataset and $L_{max}$ values.}
    \label{fig:similarity_scores_linux}
\end{figure*}

\begin{figure*}[t]
    \centering
    \includegraphics[width=1.8\columnwidth]{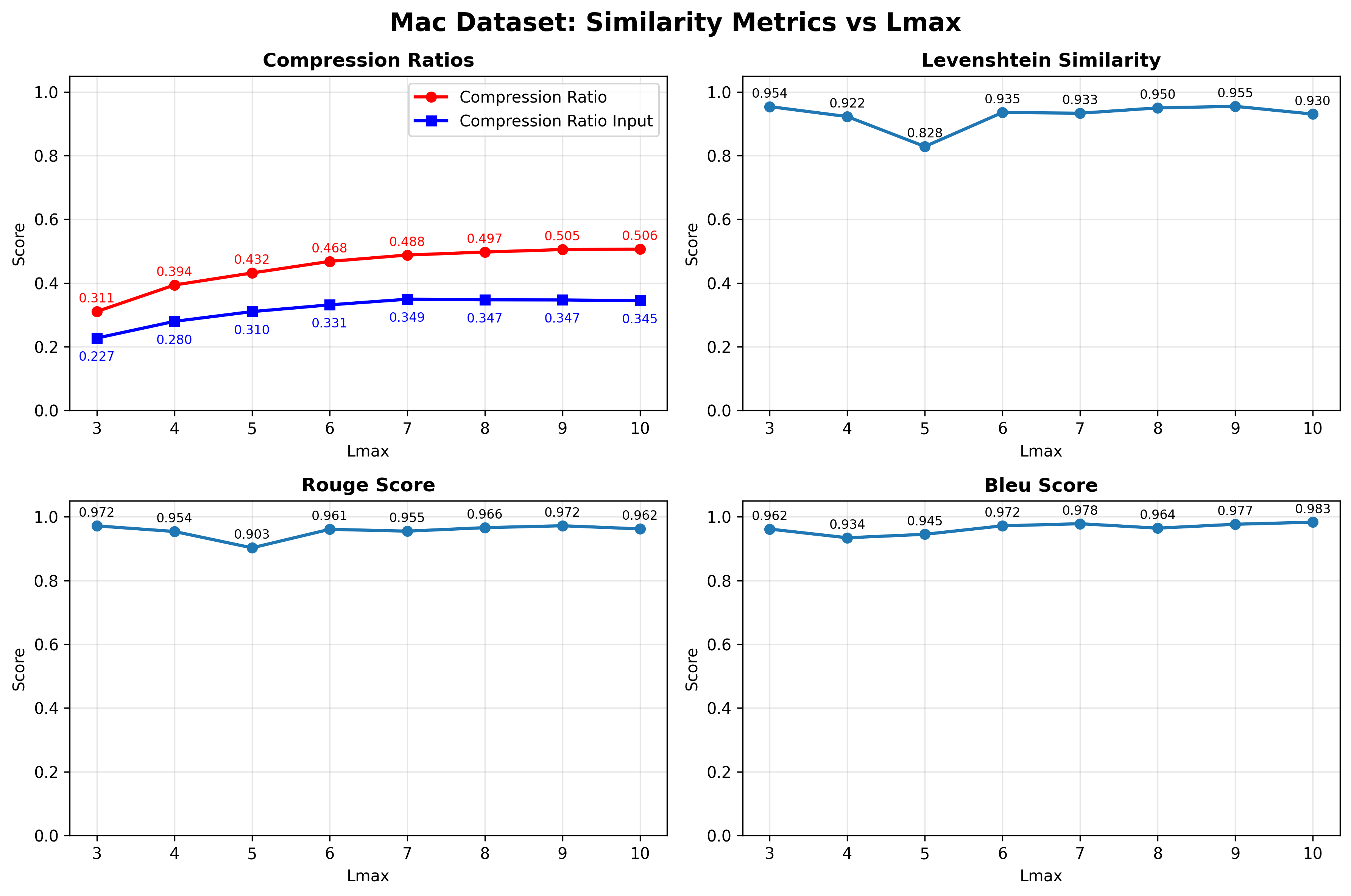}
    \caption{Levenshtein similarity, ROUGE, and BLEU scores for algorithmic compression for Mac dataset and $L_{max}$ values.}
    \label{fig:similarity_scores_mac}
\end{figure*}

\begin{figure*}[t]
    \centering
    \includegraphics[width=1.8\columnwidth]{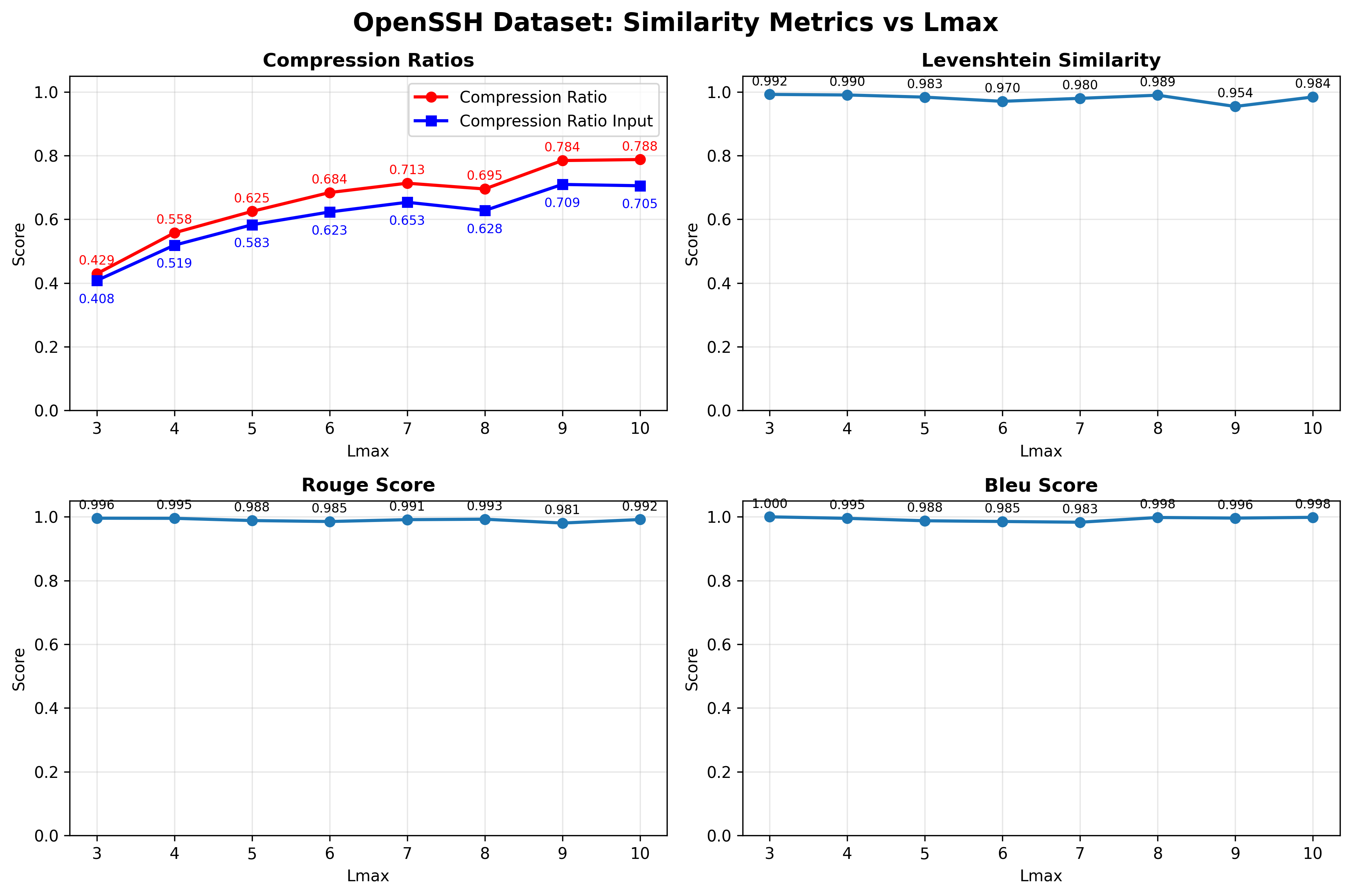}
    \caption{Levenshtein similarity, ROUGE, and BLEU scores for algorithmic compression for OpenSSH dataset and $L_{max}$ values.}
    \label{fig:similarity_scores_openssh}
\end{figure*}

\begin{figure*}[t]
    \centering
    \includegraphics[width=1.8\columnwidth]{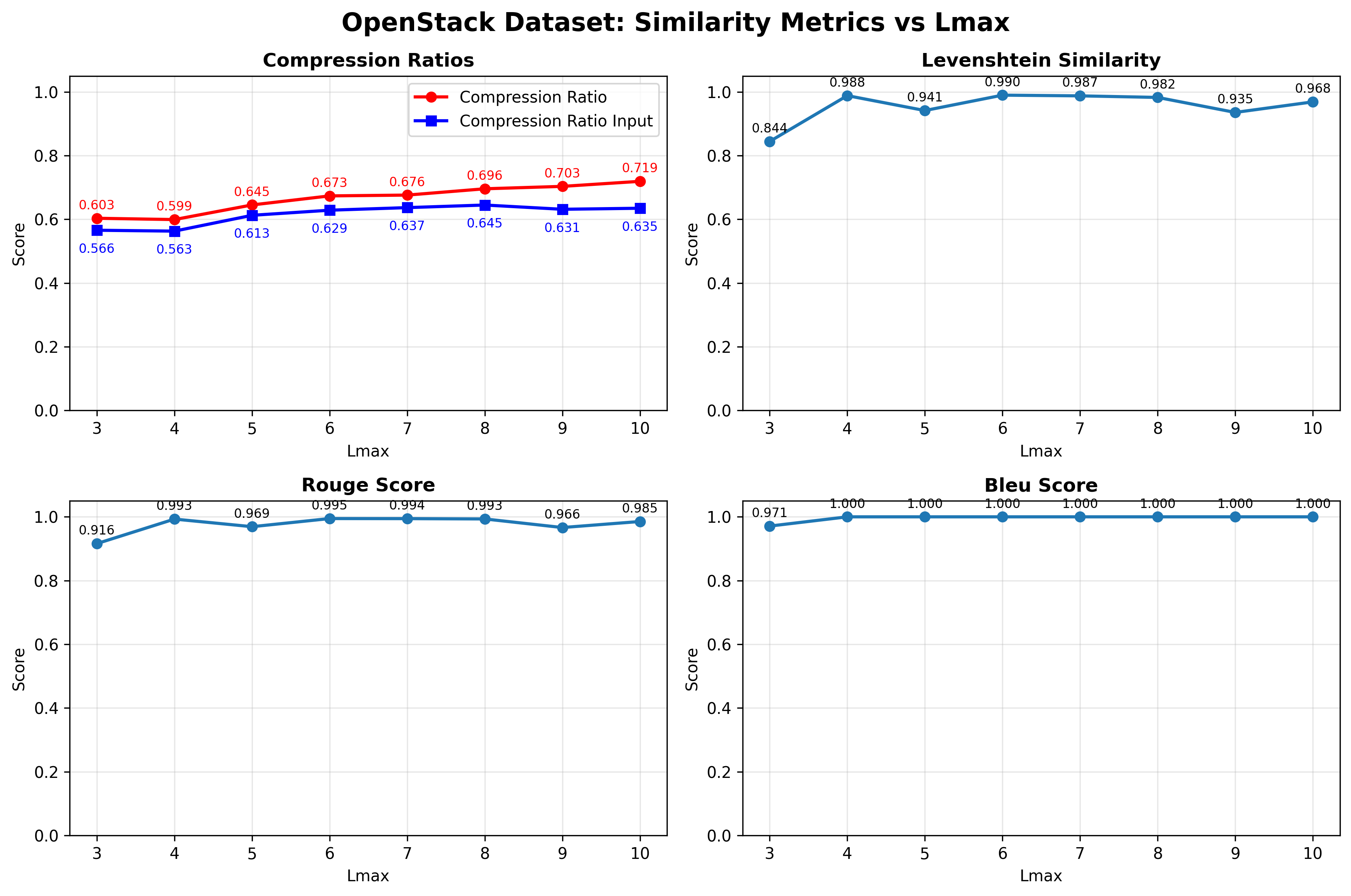}
    \caption{Levenshtein similarity, ROUGE, and BLEU scores for algorithmic compression for OpenStack dataset and $L_{max}$ values.}
    \label{fig:similarity_scores_openstack}
\end{figure*}

\begin{figure*}[t]
    \centering
    \includegraphics[width=1.8\columnwidth]{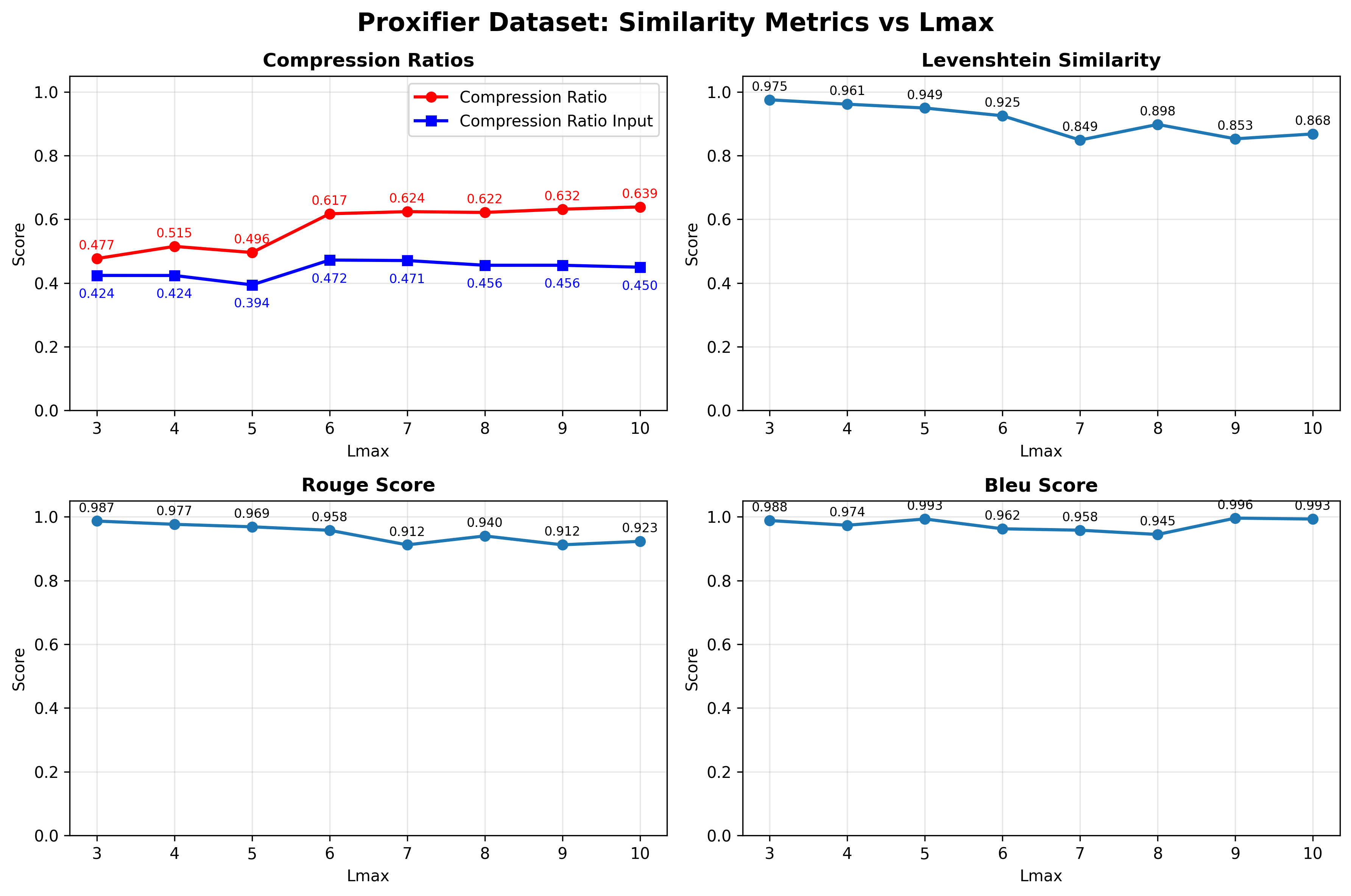}
    \caption{Levenshtein similarity, ROUGE, and BLEU scores for algorithmic compression for Proxifier dataset and $L_{max}$ values.}
    \label{fig:similarity_scores_proxifier}
\end{figure*}

\begin{figure*}[t]
    \centering
    \includegraphics[width=1.8\columnwidth]{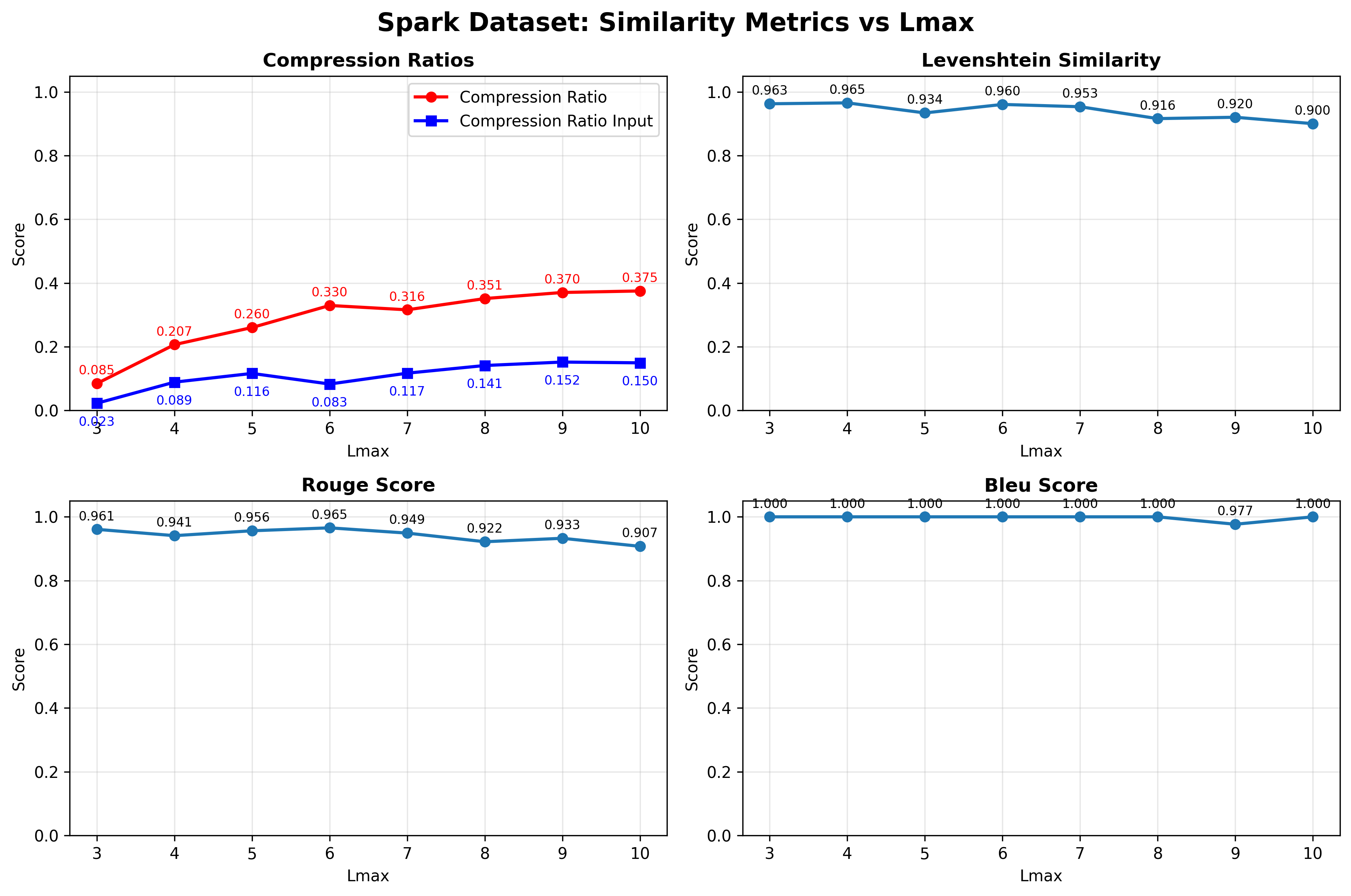}
    \caption{Levenshtein similarity, ROUGE, and BLEU scores for algorithmic compression for Spark dataset and $L_{max}$ values.}
    \label{fig:similarity_scores_spark}
\end{figure*}

\begin{figure*}[t]
    \centering
    \includegraphics[width=1.8\columnwidth]{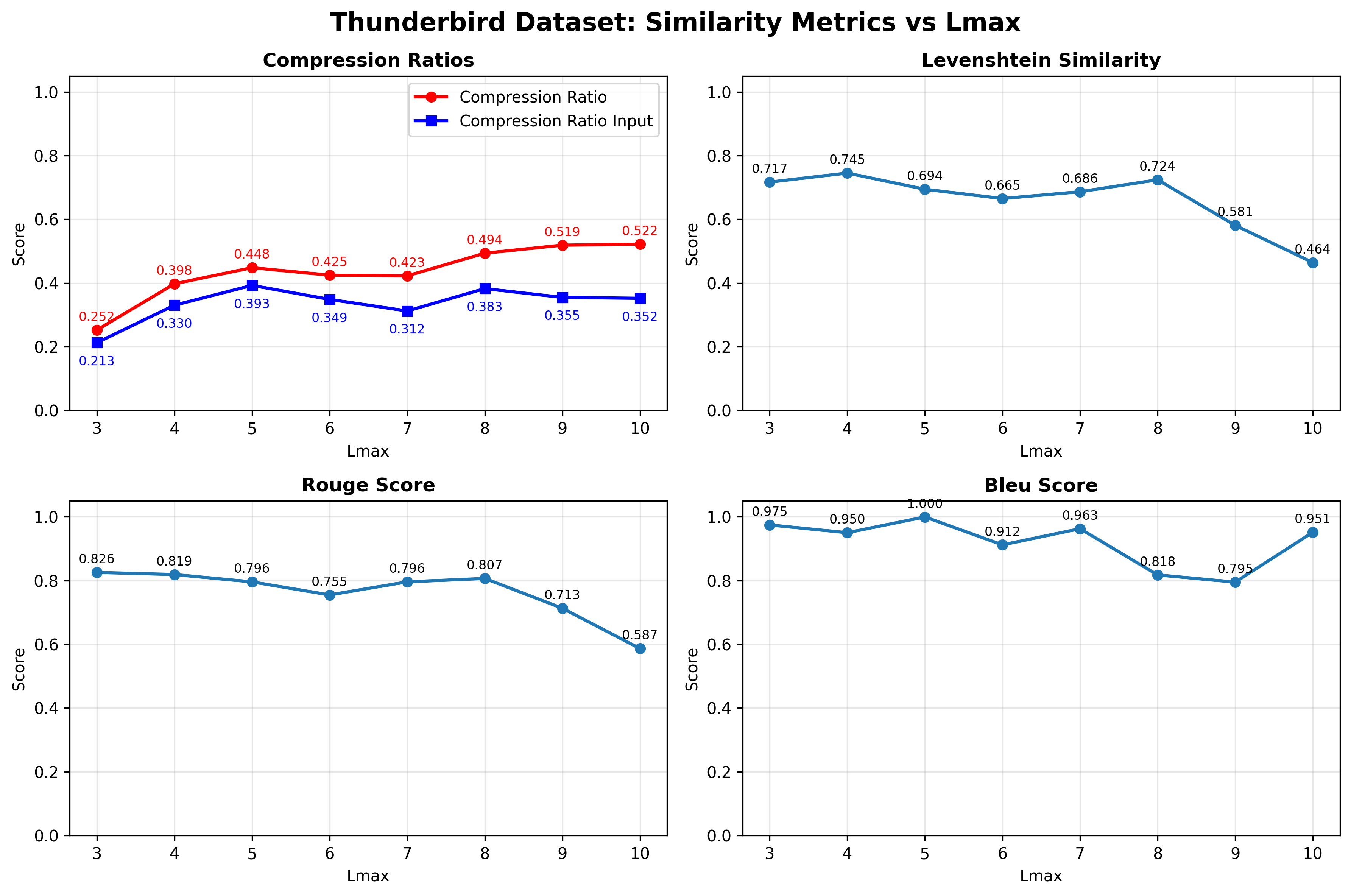}
    \caption{Levenshtein similarity, ROUGE, and BLEU scores for algorithmic compression for Thunderbird dataset and $L_{max}$ values.}
    \label{fig:similarity_scores_thunderbird}
\end{figure*}

\begin{figure*}[t]
    \centering
    \includegraphics[width=1.8\columnwidth]{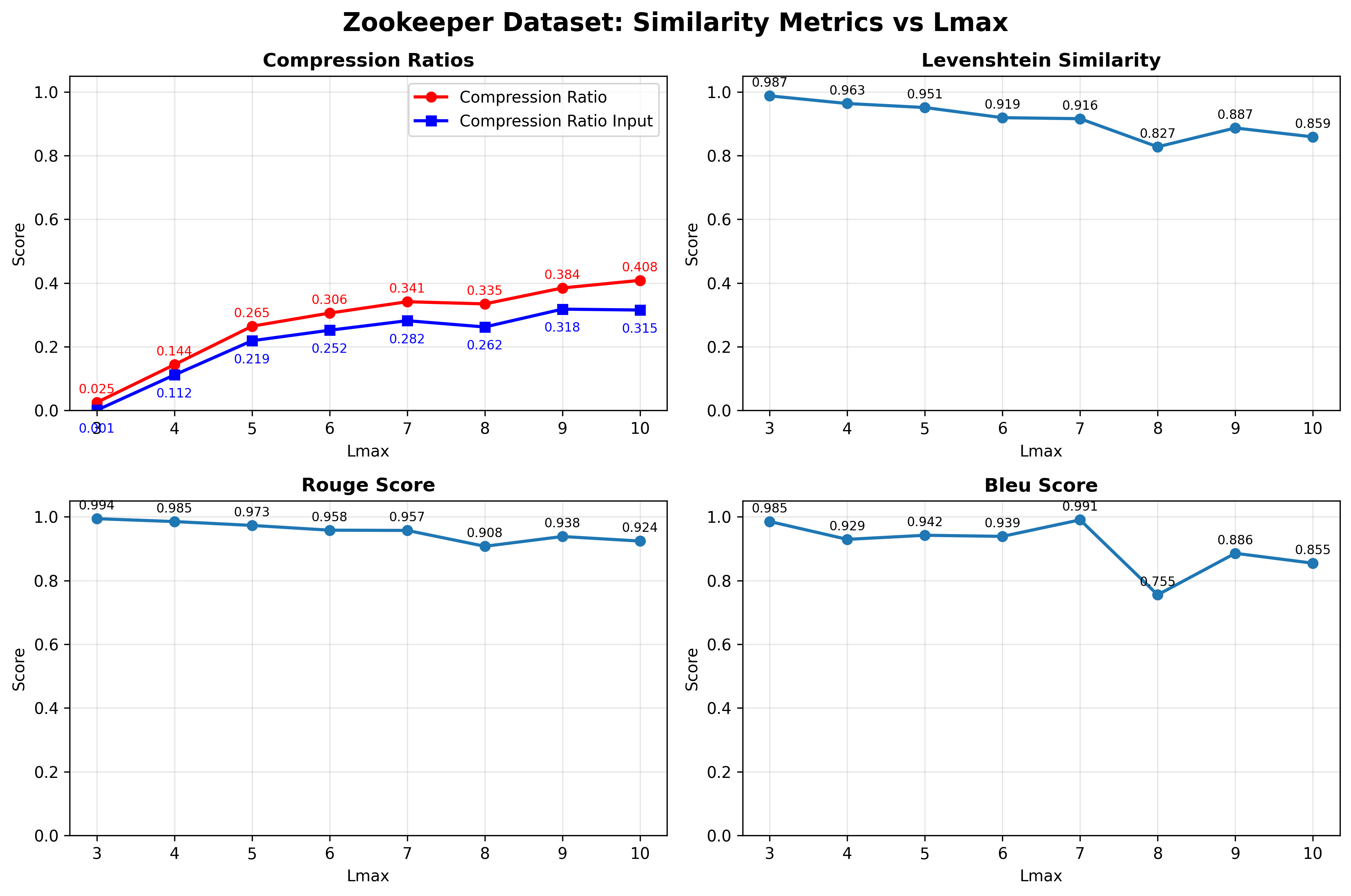}
    \caption{Levenshtein similarity, ROUGE, and BLEU scores for algorithmic compression for Zookeeper dataset and $L_{max}$ values.}
    \label{fig:similarity_scores_zookeeper}
\end{figure*}

Figures~\ref{fig:similarity_scores_apache}--\ref{fig:similarity_scores_zookeeper} present per-dataset similarity metrics (Levenshtein, ROUGE, BLEU) as a function of $L_{max}$ for the algorithmic compression experiments. Each figure shows results for a single dataset across compression levels.

\clearpage

\clearpage

\clearpage

\clearpage

\clearpage

\clearpage

\end{document}